\documentclass[journal]{IEEEtran}

\usepackage{graphicx} 
\usepackage{lineno}
\usepackage[colorlinks,
linkcolor=red,
anchorcolor=red
citecolor=red,
]{hyperref}
\usepackage{bm}
\usepackage{graphics}
\usepackage{subfig}
\usepackage{float}
\usepackage[ruled,linesnumbered]{algorithm2e}
\usepackage{algorithmicx}
\usepackage{algpseudocode}
\usepackage{setspace}
\usepackage{amsmath,amssymb,amsfonts}
\usepackage{enumitem}
\usepackage{array}
\usepackage{epstopdf}
\usepackage{mathtools}
\usepackage{caption}
\usepackage{graphicx, subfig}
\usepackage{chngcntr}
\usepackage{mathrsfs}
\usepackage{makecell,multirow,diagbox}
\usepackage{bm}
\usepackage{colortbl}
\usepackage{xcolor}
\usepackage{ntheorem}
\usepackage{placeins}
\usepackage{tablefootnote}
\usepackage{threeparttable}
\usepackage{makecell}
\usepackage{subfloat}
\usepackage[numbers,sort&compress]{natbib}
\definecolor{mygreen}{rgb}{0   0.4392 	0.2}					

\def \user2{\bm}

\begin{document}
\title{SwinV2DNet: Pyramid and Self-Supervision Compounded Feature Learning for Remote Sensing Images Change Detection}

\author{Dalong~Zheng,
		Zebin~Wu,~\IEEEmembership{Senior~Member,~IEEE,}
		Jia~Liu,~\IEEEmembership{Member,~IEEE,}
        Zhihui~Wei,~\IEEEmembership{Member,~IEEE}
\thanks{
Manuscript received DD MM, YY; revised DD MM, YY; accepted DD MM, YY. Date of publication MM DD, YY. 
This work was supported in part by the National Natural Science Foundations of China under Grant 62071233, Grant 61971223, Grant 62276133, and Grant 61976117; in part by the Jiangsu Provincial Natural Science Foundations of China under Grant BK20211570, Grant BK20180018, and Grant BK20191409; in part by the Fundamental Research Funds for the Central Universities under Grant 30917015104, Grant 30919011103, Grant 30919011402, and Grant 30921011209; in part by the Key Projects of University Natural Science Fund of Jiangsu Province under Grant 19KJA360001; and in part by the Qinglan Project of Jiangsu Universities under Grant D202062032.
(\textit{Corresponding~author:~Zebin~Wu.})
}
\thanks{\IEEEcompsocthanksitem The authors are with the School of Computer Science and Engineering, Nanjing University of Science and Technology (NJUST), Nanjing 210094, China (e-mail: zhengdl@njust.edu.cn, wuzb@njust.edu.cn, omegaliuj@njust.edu.cn, gswei@njust.edu.cn).\protect
}
}	

\markboth{Journal of \LaTeX\ Class Files,~Vol.~, No.~, MM~YY}%
{Shell \MakeLowercase{\textit{et al.}}: Bare Demo of IEEEtran.cls for IEEE Journals}

\maketitle
\begin{abstract}
Among the current mainstream change detection networks, transformer is deficient in the ability to capture accurate low-level details, while convolutional neural network (CNN) is wanting in the capacity to understand global information and establish remote spatial relationships. Meanwhile, both of the widely used early fusion and late fusion frameworks are not able to well learn complete change features. Therefore, based on swin transformer V2 (Swin V2) and VGG16, we propose an end-to-end compounded dense network SwinV2DNet to inherit the advantages of both transformer and CNN and overcome the shortcomings of existing networks in feature learning. Firstly, it captures the change relationship features through the densely connected Swin V2 backbone, and provides the low-level pre-changed and post-changed features through a CNN branch. Based on these three change features, we accomplish accurate change detection results. Secondly, combined with transformer and CNN, we propose mixed feature pyramid (MFP) which provides inter-layer interaction information and intra-layer multi-scale information for complete feature learning. MFP is a plug and play module which is experimentally proven to be also effective in other change detection networks. Further more, we impose a self-supervision strategy to guide a new CNN branch, which solves the untrainable problem of the CNN branch and provides the semantic change information for the features of encoder. The state-of-the-art (SOTA) change detection scores and fine-grained change maps were obtained compared with other advanced methods on four commonly used public remote sensing datasets. The code is available at {\url{https://github.com/DalongZ/SwinV2DNet}}.
\end{abstract}


\IEEEpeerreviewmaketitle

\section{Introduction}
\begin{figure}[t]
	\centering
	\includegraphics[width=0.99\linewidth]{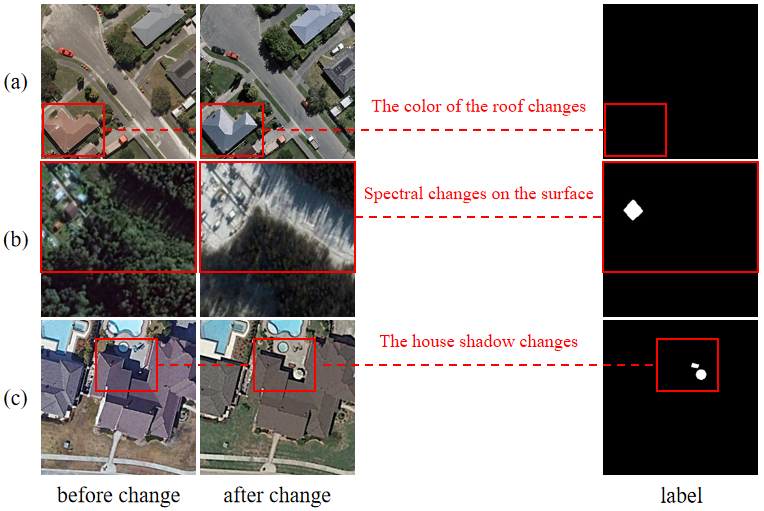}
	\caption{A variety of pseudo changes become the challenges in change detection. (a) The roof color changes. (b) Spectral changes on the surface. (c) The house shadow changes.}
	\label{fig1}
\end{figure}

\IEEEPARstart{R}{emote} sensing images change detection is one of the earliest and most important remote sensing tasks, which has been concerned and studied by many researchers for a long time \cite{hussain2013change,zhang2021moving,bruzzone2009domain,benedek2011building}. Change detection is defined as observing the image differences of the same surface area at different times. Change detection is used in many scenarios, including disaster assessment \cite{lu2019landslide}, urban planning, land surface change \cite{jin2017land,zhu2014continuous}, and so on.
With the development of satellites and sensors, very high resolution remote sensing (VHRRS) images have gradually become one of the mainstream remote sensing images in research, which provide rich spatial information and fine surface details. However, one of the main challenges faced by VHRRS images change detection is high intraclass variation and low interclass variance of detection objects \cite{lv2021land}. It, therefore, has been the focus of scholars' research that how to design a stable network and provide comprehensive and diverse feature information to distinguish the pseudo changes in change detection (as shown in Fig. \ref{fig1}).

Traditional change detection algorithms, according to different detection units, can be divided into pixel-based algorithms and object-based algorithms. The detection results of pixel-based algorithms are obtained through feature extraction and then threshold segmentation, which include methods based on arithmetic operations (band difference \cite{liangpei2017advance}, spectral angle mapper \cite{zhuang2016strategies}), methods based on transformation (change vector analysis (CVA) \cite{malila1980change,bovolo2006theoretical}, principal component analysis (PCA) \cite{zhang2007remote}, independent component analysis (ICA) \cite{zhong2006multi}), post-classification change detection \cite{xian2010updating}, slow feature analysis (SFA) \cite{wu2013slow} and so on. According to the shape, texture and spectrum of images, object-based algorithms need to segment the images and then compare the classification results to get the change detection results \cite{gil2016description}. Pixel-based algorithms are trapped by the interference of small noises and the decision of segmentation threshold. Meanwhile object-based algorithms often get stuck in the accumulation of multiple classification errors that affect the detection accuracy \cite{hussain2013change}. Both of these traditional algorithms require prior knowledge and manual design, and are easily affected by sensor noises.

With the support of massive remote sensing data, deep learning has also shown outstanding detection ability in the field of remote sensing. CNN converts the input images into the high-dimensional depth features, and combines the targets and background to extract effective semantic information, achieving the detection effect beyond many traditional methods. \cite{daudt2018fully} provides the three most common baseline networks for change detection. The architecture combined with CNN and conditional random field (CRF) refines the edges of detection areas, but it comes at the cost of slow training \cite{zheng2022learning}. CNN is hindered regrettably by the narrow receptive field of local information, and transformer rises rapidly due to the ability of modeling global information. However, it can not work well that the pure transformer change detection model lacks low-level details \cite{zhang2022swinsunet}. Therefore, how to combine transformer and CNN to build the reasonable change detection architecture is the crux of the matter at this stage.

From another perspective, change detection network architectures can be divided into early fusion (EF) \cite{daudt2018fully,alcantarilla2018street,peng2019end,peng2020optical} and late fusion (LF) \cite{daudt2018fully,zhan2017change,zhang2020deeply,chen2021remote,shi2021deeply,fang2021snunet,zhang2023asymmetric,feng2023change} networks. The EF network works by stitching two images together and feeding them into the single input network. By concatenating two three-channel images into a six-channel image, \cite{alcantarilla2018street} and \cite{peng2019end} input it into full convolutional neural network (FCN) and UNet++ respectively, and output change map after training the network. The disadvantage of this method is that the network lacks the depth features of single images, resulting in fractured edges and broken structures in change map. 
In the LF network, the two features are extracted from the pre-changed and post-changed images respectively by using the dual-input structure, and are fused in the second half of the network. The siamese network, the most prominent LF network, consists of two subnets with shared weights. The siamese network was first used for remote sensing images change detection in \cite{zhan2017change}. The use of convolutional block attention module (CBAM) and deep supervision for the siamese network respectively alleviates the problem of heterogeneous features fusion and depth features migration in training process \cite{zhang2020deeply}. However, for the LF network, the contradiction between the dual-stream input of encoder and the single output of decoder often results in the disappearance of gradient propagation and affects the low-level features learning of two original images. Furthermore, the heterogeneous features fusion of LF network needs to be solved by elaborately designed module. 
As a consequence, it is another problem worth pondering that how to overcome the respective disadvantages of these two network architectures and provide the complete and diverse features for change detection.

In addition to the design of the overall network architectures for change detection, researchers are also pushing forward the elaboration of network functional modules. The attention modules introduced into change detection are relatively representative, including squeeze-and-excitation attention (SE) \cite{hu2018squeeze}, efficient channel attention (ECA) \cite{wang2020eca}, CBAM \cite{woo2018cbam} and cross-attention \cite{zhang2023asymmetric}. At the same time, since the ground objects have different scales in the VHRRS change detection datasets, how to adapt these ground objects to maintain the robustness and generalization ability of the network? Multi-scale features of deep learning generally can be divided into three categories: multi-scale features between different layers, multi-scale interaction features between different layers and multi-scale features from different convolution units. The first type of multi-scale features was embedded in the common U-Net network. The second type of multi-scale features typically interacts and fuses using transformer or CNN. The third type of multi-scale features is provided by a variety of convolution units, such as inception \cite{szegedy2015going}, dilated convolution \cite{yu2015multi}, res2net convolution (Res2Net-Conv) \cite{gao2019res2net}, selective kernel convolution (SK-Conv) \cite{li2019selective} and so on.
We should think about the integration and utilization of these three multi-scale features. Other scholars also think in the combination with generative adversarial network (GAN) \cite{hou2019w,zhao2019incorporating} or self-supervised learning \cite{chen2022fccdn,9874899,10109662} to obtain more discriminative features. These deep learning technologies are aimed at solving the problem of high intraclass variation and low interclass variance by mining the different features of change detection data.

Motivated by the above concerns, this study combines Swin V2 and VGG16 to propose a new end-to-end compounded dense network SwinV2DNet. Swin V2 blocks are used to build the UNet++ type main network, and VGG16 encoder is used to build the CNN auxiliary network. SwinV2DNet overcomes the modeling defect of only local information in CNN and the insufficient interpretation of low-level details in transformer. On the other hand, the Swin V2 main network belongs to EF network, and the CNN branch belongs to LF. This structure constantly provides the pre-changed features, post-changed features and change relation features (namely, the six-channel concatenation from the pre-changed and post-changed images) for the accurate acquisition of change detection results. 
CBAM and deep supervision also promote the fusion of heterogeneous features and the rapidly stable convergence of the network, respectively. 
To better combine transformer and CNN, we propose a new multi-scale module, mixed feature pyramid, which provides inter-layer multi-scale interaction information and intra-layer multi-scale information to supplement the UNet++ main network only with inter-layer multi-scale information. 
We finally design a new decoder to the CNN branch with only VGG16 encoder, and use the self-supervision strategy to train the extracted features, so that the CNN branch can provide learnable and more discriminant semantic information. 
To sum up, the main contributions of this study are fourfold:

\begin{itemize}
	\item[1)] We propose an end-to-end compounded dense network SwinV2DNet that possesses both advantages of transformer and CNN, and overcomes respective disadvantages of the EF and LF network. This is the first parallel combination of Swin V2 and VGG16 in change detection. 
 	\item[2)] Mixed feature pyramid is proposed, for the first time, to provide inter-layer interaction information and intra-layer multi-scale information. It is a plug and play module that has been experimentally proven to be also effective in other change detection networks. 
	\item[3)] We design a new decoder for the CNN branch with only VGG16 encoder, and impose the self-supervised strategy to train the extracted features to provide more discriminative semantic information for the main network.
	\item[4)] Compared with other advanced methods, our method obtains the state-of-the-art (SOTA) change detection scores and the elaborate change maps on four common public remote sensing datasets. 
\end{itemize}

The remainder of this paper is organized as follows. 
Section \ref{section2} reviews the related work.  
Section \ref{section3} elaborates the proposed SwinV2DNet method.
The experimental evaluations and ablation studies are carried out in Section \ref{section4}. 
Finally, Section \ref{section5} presents the conclusion of this article.

\section{Related Work}
\label{section2}
\subsection{Visual Transformer}
Transformer first has been stand out in natural language processing (NLP) due to its ability of modeling global information \cite{vaswani2017attention}. Vision transformer (ViT) is proposed by \cite{dosovitskiy2020image} to bring the spirit of self-attention to image classification, but this huge model cannot be directly applied to other computer vision detection tasks. Swin transformer adopts the inspiration of architectural hierarchy refinement and local information interaction, which reduces the heavy computation involved in modeling global information \cite{liu2021swin}. Furthermore, \cite{liu2022swin} proposes Swin V2 that stabilizes the training process caused by the increase of model parameters and mitigates the resolution difference between the upstream and downstream tasks. At the same time, various transformer models are developing rapidly in the field of remote sensing images processing \cite{hong2021spectralformer,sun2022spectral,gao2023adaptive,he2022swin}.

Due to the lack of low-level details in the pure transformer model \cite{zhang2022swinsunet}, transformer is generally combined with CNN for change detection. On the one hand, in the serial combination methods, Chen et al. proposes bitemporal image transformer (BiT) that embeds transformer to CNN to enhance the ability of modeling contexts within the spatial-temporal
domain \cite{chen2021remote}. TransUNetCD mades up for the former omission of the high-resolution shallow information by cascading the up-sampling decoder to achieve the more detailed change map \cite{li2022transunetcd}. On the other hand, in the parallel associative approachs, \cite{feng2022icif} uses transformer and CNN respectively to extract the pair change features for feature fusion, but it also ignores the supplement of low-level details at the decoder stage. Our study adopts a new parallel approach with the high and low level features association to avoid the defects of the above models and obtain the best detection result.

\subsection{Multi-Scale Information}
\label{section2_2}
Multi-scale information is always an important tool in image processing. In deep learning, multi-scale features can be divided into three categories: multi-scale features between different layers, multi-scale interaction features between different layers and multi-scale features from different convolution units. The first category is often already implicit in a variety of classic networks, such as ResNet, FCN and U-Net \cite{He2015DeepRL,Shelhamer2014FullyCN,DBLP:journals/corr/RonnebergerFB15}. Feature pyramid transformer (FPT) uses two different transformers to interfuse the features between different layers to generate the second multi-scale features \cite{Zhang2020FeaturePT}. The third type of multi-scale features is provided by various convolution units, such as inception, dilated convolution, Res2Net-Conv, SK-Conv and so on.

In VHRRS images change detection, \cite{Feng2023ChangeDO} extracts the multilevel intertemporal features through the double branches of shared weights, and then performs the information fusion and features difference to obtain the robust change features. The multiscale decoupled convolution is constructed by using atrous convolutions with different dilation rates. The researchers embed several of these convolutions in different layers to acquire the two multiscale features between and within layers \cite{Lei2023UltralightweightSF}. Based on FPT, we propose mixed feature pyramid that provides inter-layer multi-scale interaction information and intra-layer multi-scale information to supplement the main network only with inter-layer multi-scale information.

\subsection{Self-Supervised Learning}
\label{Section2_3}
Since the labeled data requires precious labor costs, self-supervised learning (SSL), in which a large amount of unlabeled data can be used to train networks and extract knowledge that are then transferred to downstream tasks, has become a flourishing deep learning technology recently. SSL is first used in NLP. \cite{devlin2018bert} designs the automatic regression task for words, which provides more contextual information than the task of matching a single sentence to a single label. Then SSL is introduced into computer vision. Contrast learning is one of the representative models \cite{chen2020simple}, that is, the image sample pairs are generated through the network, and the cost function is set to shorten the distance between the two positive samples as far as possible and expand the distance between the positive and negative samples. Mask autoencoder (MAE) is another important branch of SSL \cite{he2022masked}.

In change detection, Chen and Bruzzone \cite{chen2103self} use bootstrap your own latent (BYOL) framework to pretrain the heterogeneous remote sensing images, and then transfer it to the downstream change detection task. The use of ViT encoder and random masking as SSL data enhancement techniques further advances this architecture \cite{zhang2023self}. SSL has been shown to provide the more discriminative semantic information when combined with the supervised techniques to train the network \cite{chen2022fccdn}. Different from \cite{chen2022fccdn}, we design a new encoder-decoder architecture with VGG16 as the encoder, and apply SSL to this branch architecture to provide the self-supervised semantic features for the main network.

\section{Proposed Method}
\label{section3}
In this section, we first introduce SwinV2DNet architecture. Moreover, Swin V2 block, MFP module, the CNN branch network and SSL strategy are described respectively. Finally, we specify the loss function of the model.

\begin{figure*}[t]
	\setlength{\belowcaptionskip}{-0.1cm}
	\centering
	\includegraphics[width=0.99\linewidth]{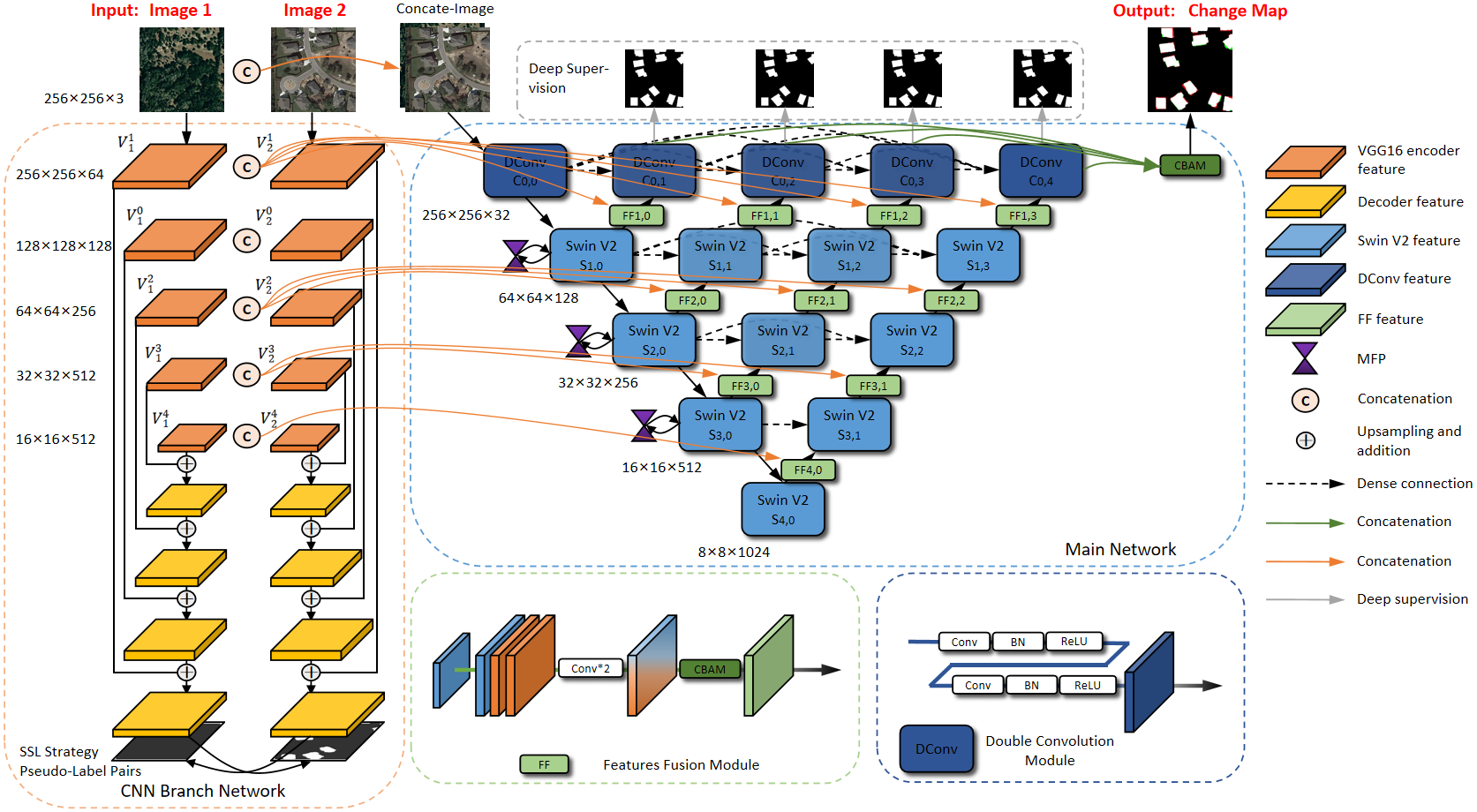}
	\caption{The overall architecture of SwinV2DNet.}
	\label{fig2}
	\vspace{-0.3cm}
\end{figure*}
\subsection{Proposed SwinV2DNet}
\label{section3_1}
The overall architecture of SwinV2DNet is shown in Fig. \ref{fig2}. In conjunction with Algorithm \ref{alg1}, we elaborate on the details of the network. First of all, we initialize the parameters ${i}$ and ${j}$ to satisfy the following conditions:
\begin{equation}
	\vspace{-0.1cm}
	\label{equation1}
	\begin{array}{l}
		1 \le i \le 4,i \in N\\
		0 \le j \le 3,j \in N\\
		whenj = 0,i \in \{ 1,2,3,4\} \\
		whenj = 1,i \in \{ 1,2,3\} \\
		whenj = 2,i \in \{ 1,2\} \\
		whenj = 3,i \in \{ 1\} 
	\end{array}
\end{equation}
where ${N}$ is the set of natural numbers. ${i}$ is the number of layers of SwinV2DNet or VGG16, and ${j}$ is the number of columns of SwinV2DNet. Then we obtain the encoder features of CNN branch network to compensate for the lack of low-level details and single image convolution features of main network: 
\begin{equation}
	\vspace{-0.1cm}
	\label{equation2}
	V_1^{\rm{i}},V_2^{\rm{i}} = CN{N_{SSL}}(I_{1},{I_2})
\end{equation}
${CN{N_{SSL}}}$ refers to CNN branch network trained using SSL strategy. ${V_1^{\rm{i}},V_2^{\rm{i}}}$ and ${I_{1},{I_2}}$ represent the encoder feature pair and the original image pair respectively in this paper. On another path in parallel, the features in the first column of main network are generated:
\begin{equation}
	\vspace{-0.1cm}
	\label{equation3}
	{C_{0,0}},{{S'}_{i,j = 0}} = SwinV{2_{backbone}}(DConv(Cat({I_{1}},{I_2})))
\end{equation}
${SwinV{2_{backbone}}}$, ${DConv}$ and ${Cat}$ refer to Swin V2 backbone \cite{liu2022swin}, double convolution module (DConv) and the concatenation in the channel dimension, respectively. ${C}$ denotes the feature from DConv. ${S'}$ denotes the feature in column 1 from Swin V2 backbone. The features are estimated by MFP that provides inter-layer multi-scale interaction information and intra-layer multi-scale information:
\begin{equation}
	\vspace{-0.1cm}
	\label{equation4}
		{S_{1,0}},{S_{2,0}},{S_{3,0}} = MFP({{S'}_{1,0}},{{S'}_{2,0}},{{S'}_{3,0}}); 
		{S_{4,0}} = {{S'}_{4,0}}
\end{equation}
\begin{algorithm}[!t]
	\begin{small}
		\caption{Inference of SwinV2DNet Model for Change Detection}
		\label{alg1}
		\LinesNumbered
		\KwIn {image 1 ${I_{1}}$, image 2 ${I_{2}}$.}
		\KwOut{change map ${CM}$.}
		{\bf{Initialization:}} set parameters ${i}$, ${j}$ satisfying Eq. (\ref{equation1}).\\
		
		{\bf{Acquisition of encoder features for CNN branch network:}} update ${V_{1}^{i}}$, ${V_{2}^{i}}$ using ${I_{1}}$, ${I_{2}}$ via Eq. (\ref{equation2}).\\
		
		{\bf{Acquisition of features in column 1 for main network:}} update ${C_{0,0}}$, ${S'_{i,j=0}}$ using ${I_{1}}$, ${I_{2}}$ via Eq. (\ref{equation3}).\\
		
		{\bf{Estimation of MFP features:}} update ${S_{1,0}}$, ${S_{2,0}}$, ${S_{3,0}}$ and ${S_{4,0}}$ by solving Eq. ({\ref{equation4}}).\\
		{\bf{Acquisition of Swin V2 and FF features in rows 1 to 4 for main network:}}\\
		\For{$j=0$ \KwTo $3$}{
			\uIf{$j==1$}{
				compute ${S_{i,1}}$ by ${S_{i,0}}$, ${FF_{i+1,0}}$ via Eq. (\ref{equation5}); \\
			}\uElseIf{$j==2$}{
				compute ${S_{i,2}}$ by ${S_{i,0}}$, ${S_{i,1}}$, ${FF_{i+1,1}}$ via Eq. (\ref{equation6}); \\
			}\ElseIf{$j==3$}{
				compute ${S_{i,3}}$ by ${S_{i,0}}$, ${S_{i,1}}$, ${S_{i,2}}$ and ${FF_{i+1,2}}$ via Eq. (\ref{equation7}). \\
			}
			Update ${FF_{i,j}}$ using ${S_{i,j}}$, ${V_{1}^{i}}$, ${V_{2}^{i}}$ via Eq. (\ref{equation8}).\\
		}
		
		{\bf{Estimation of other DConv features:}} compute ${C_{0,1}}$, ${C_{0,2}}$, ${C_{0,3}}$ and ${C_{0,4}}$ by solving Eq. ({\ref{equation9}}).\\
		
		{\bf{Generation of change map:}} update ${CM}$ via Eq. (\ref{equation10}).   
	\end{small}     
\end{algorithm}
${S}$ refers to the feature that comes from MFP or Swin V2 block. 

At this point, the rest of Swin V2 and FF features in rows 1 to 4 of main network can be generated:
\begin{equation}
	\vspace{-0.1cm}
	\label{equation5}
	{S_{i,1}} = SwinV2(Conv(Cat({S_{i,0}},Up(F{F_{i + 1,0}}))))
\end{equation}
\begin{equation}
	\vspace{-0.5cm}
	\label{equation6}
	{S_{i,2}} = SwinV2(Conv(Cat({S_{i,0}},{S_{i,1}},Up(F{F_{i + 1,1}}))))
\end{equation}
\begin{equation}
	\vspace{-0.5cm}
	\label{equation7}
	{S_{i,3}} = SwinV2(Conv(Cat({S_{i,0}},{S_{i,1}},{S_{i,2}},Up(F{F_{i + 1,2}}))))
\end{equation}
\begin{equation}
	\vspace{-0.1cm}
	\label{equation8}
	F{F_{i,j}} = FF({S_{i,j}},V_1^{\rm{i}},V_2^{\rm{i}})
\end{equation}
${FF()}$ represents feature fusion module (FF), and ${FF}$ represents the FF features.
${SwinV2}$ represents Swin V2 block.
${Conv}$ and ${Up}$ are used to adjust the spatial and channel resolutions of the features.
Then other DConv features are acquired:
\begin{equation}
	\vspace{-0.1cm}
	\label{equation9}
	\begin{array}{l}
		{C_{0,1}} = DConv(Cat({C_{0,0}},F{F_{1,0}}))\\
		{C_{0,2}} = DConv(Cat({C_{0,0}},{C_{0,1}},F{F_{1,1}}))\\
		{C_{0,3}} = DConv(Cat({C_{0,0}},{C_{0,1}},{C_{0,2}},F{F_{1,2}}))\\
		{C_{0,4}} = DConv(Cat({C_{0,0}},{C_{0,1}},{C_{0,2}},{C_{0,3}},F{F_{1,3}}))
	\end{array}
\end{equation}
Finally, we generate change map ${CM}$ via CBAM module \cite{woo2018cbam} and sigmoid function:
\begin{equation}
	\vspace{-0.1cm}
	\label{equation10}
	CM = Sig(Con{v_{1 \times 1}}(CBAM(Cat({C_{0,1}},{C_{0,2}},{C_{0,3}},{C_{0,4}}))))
\end{equation}
It should also be noted that dense connectivity allows the network to provide more diverse features. And deep supervision solves the problem of features shift in large network during training. The joint of these two techniques allows the network to train stably and converge quickly. FF concatenates, nonlinearizes, and finally uses CBAM to process the heterogeneous features for effective fusion. DConv is the double union of convolution, batch normalization (BN) and ReLU function.

To summarize, the superior detection performance of SwinV2DNet, a parallel compounded architecture, can be attributed to the following reasons. Firstly, main network consisting of Swin V2 blocks is responsible for extracting the change relationship features which are crucial for the change detection task. Secondly, CNN branch network complements main network by providing the low-level pre-changed and post-changed features necessary for accurate detection. Moreover, the incorporation of the multi-scale features and SSL semantic features further enhances the overall feature learning of the network. These combined factors contribute to SwinV2DNet achieving the SOTA detection performance.

\subsection{Swin Transformer V2 Block}
\label{section3_2}
\begin{figure}[h]
	\vspace{-0.3cm}
	\setlength{\belowcaptionskip}{-0.2cm}
	\centering
	\includegraphics[width=0.99\linewidth]{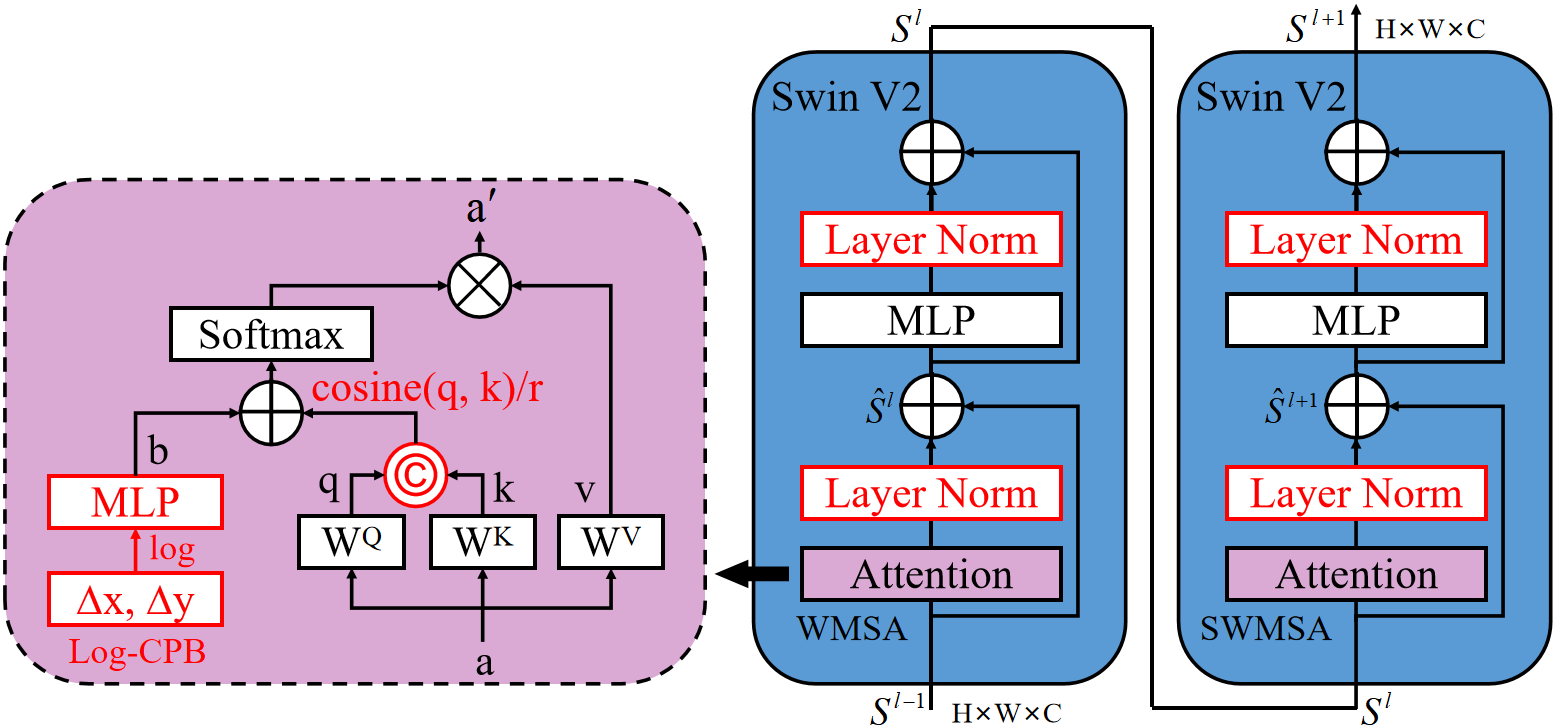}
	\caption{Swin V2 block. The improvements of Swin V2 compared to Swin V1 are marked in red.
	}
	\label{fig3}
\end{figure}

Swin-type transformers greatly reduce the number of model parameters while modeling global information through shifted window and hierarchical mechanism \cite{liu2021swin}. Swin V2 \cite{liu2022swin} further employs the post-normalization and scaled cosine attention techniques to improve the stability of the large vision model. At the same time, the  log-spaced continuous position bias method is used to alleviate the problem of transferring the model trained on low-resolution images to high-resolution images. So we use Swin V2 as the base block to build the main network. It is shown in Fig. \ref{fig3}.
\begin{figure*}[t]
	\setlength{\belowcaptionskip}{-0.1cm}
	\centering
	\includegraphics[width=0.99\linewidth]{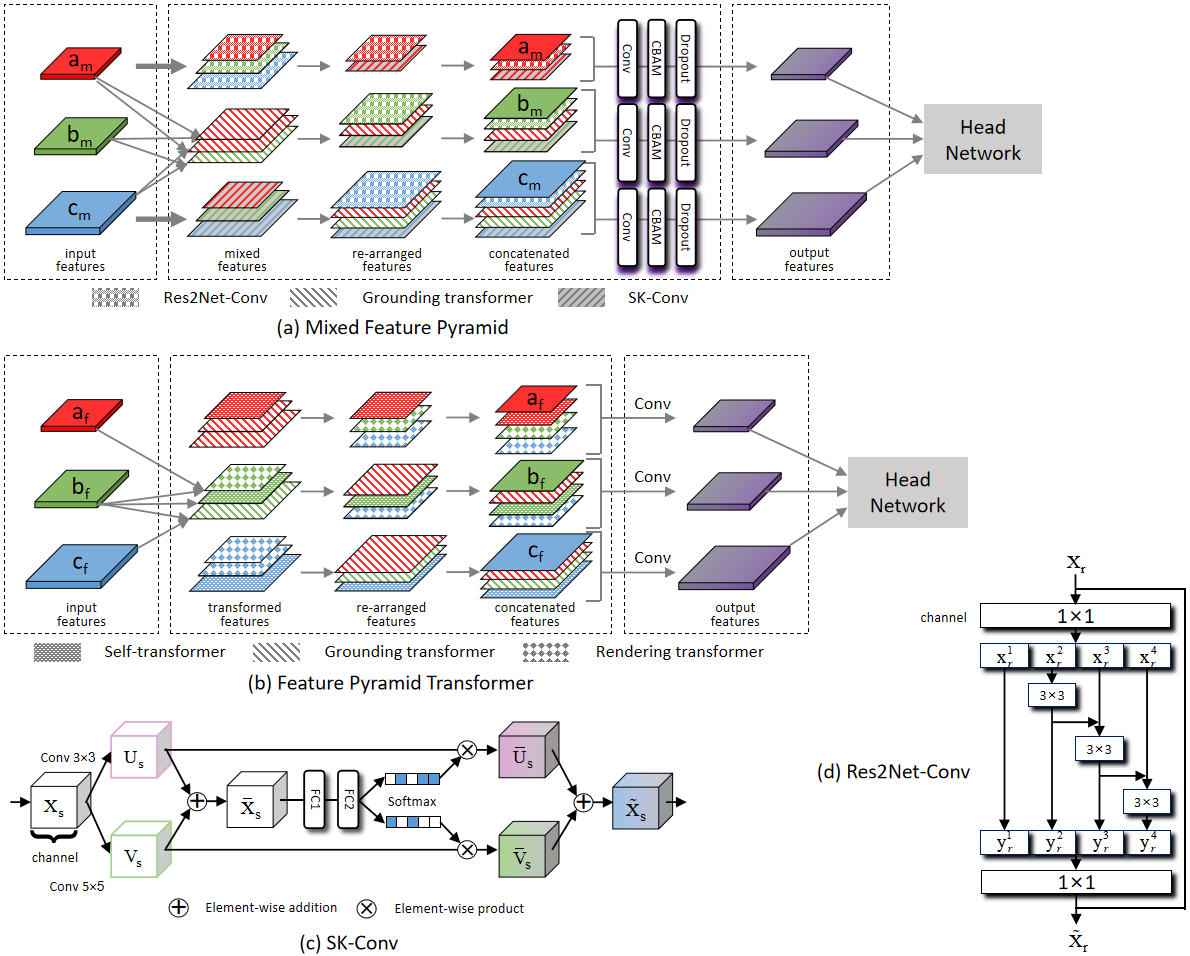}
	\caption{(a) Mixed feature pyramid. (b) Feature pyramid transformer. Mixed feature pyramid is an improvement based on feature pyramid transformer. (c) SK-Conv. (d) Res2Net-Conv. Both SK-Conv and Res2Net-Conv are the components of mixed feature pyramid.}
	\label{fig4}
\end{figure*}

Swin V2 splits the image into the patchs and then models the patchs to generate the features with global information. It is generally pairwise. Its input and output are ${S^{l-1}}$ and ${S^{l+1}}$, respectively. We place layer normalization (LN) after multi-head self-attention (MSA) or multilayer perceptron (MLP) to improve the stability of the network. MLP is to enhance the nonlinear ability of the block. MSA includes window MSA (WMSA) and shifted window MSA (SWMSA): the former extracts the image features in the windows to reduce the computational cost, and the latter maintains the interaction of global information by sliding windows. The residual connection ensures the effective dissemination of information in the large network. The specific cooperation of these components is as follows:
\begin{equation}
	\vspace{-0.1cm}
	\label{equation11}
	\begin{array}{l}
		{{\hat S}^l} = WMSA(LN({S^{l - 1}})) + {S^{l - 1}}\\
		{S^l} = MLP(LN({{\hat S}^l})) + {{\hat S}^l}\\
		{{\hat S}^{l + 1}} = SWMSA(LN({S^l})) + {S^l}\\
		{S^{l + 1}} = MLP(LN({{\hat S}^{l + 1}})) + {{\hat S}^{l + 1}}
	\end{array}
\end{equation}
Here, ${S}$ is the image feature and ${l}$ is the number of Swin V2 block layers. 

Attention is the core of the whole block. The image features are firstly mapped into the three vectors: query (Q), key (K), and value (V). Then we use Sim function (namely Eq. (\ref{equation14})) to calculate the correlation weight matrix coefficients of Q and K, and normalize these weight matrix by softmax. Finally, the dot product between the weight coefficients and V is finished to form the self-attention features:
\begin{equation}
	\vspace{-0.1cm}
	\label{equation12}
	Attention(Q,K,V) = Softmax(Sim(Q,K))V
\end{equation}
where ${Q}$, ${K}$ and ${V}$ are N${\times}$d matrices. N and d severally represent the number of patches and the dimension of single head self-attention. Instead of applying a single head self-attention, MSA of Swin V2 computes each head self-attention separately and concatenates these head self-attentions that represent different subspaces:
\begin{equation}
	\vspace{-0.1cm}
	\label{equation13}
		\begin{array}{l}
			MSA(Q,K,V) = Cat(hea{d_1},...,hea{d_H}){W^O}\\
			{\rm{\quad}}where{\rm{\quad}}hea{d_p} = Attention(QW_p^Q,KW_p^K,VW_p^V)
		\end{array}
\end{equation}
where ${W_p^Q \in {^{D \times {d_k}}}}$, ${W_p^K \in {^{D \times {d_k}}}}$, ${W_p^V \in {^{D \times {d_v}}}}$, and ${{W^O} \in {^{h{d_v} \times D}}}$ are parameter matrices. ${h}$ and ${D}$ respectively represent the number of self-attention heads and the dimension of embedding layers. We set ${{d_k} = {d_v} = D/h}$.

Swin V2 computes the attention logit of a pixel pair ${m}$ and ${n}$ by a scaled cosine function:
\begin{equation}
	\vspace{-0.1cm}
	\label{equation14}
	Sim({q_m},{k_n}) = \cos ({q_m},{k_n})/r + {B_{mn}}
\end{equation}
Here, ${B_{mn}}$ is the relative position bias between pixel ${m}$ and ${n}$. ${r}$ is a learnable scalar, non-shared across heads and layers. Furthermore, Swin V2 proposes the log-spaced coordinates instead of the original linear-spaced ones to facilitate the transferring of the model between different resolution images. The log-spaced coordinates is then used as the input of ${\Phi}$ to generate the bias values:
\begin{equation}
	\vspace{-0.1cm}
	\label{equation15}
	\begin{array}{l}
		\widehat {\Delta x} = sign(x) \cdot \log (1 + |\Delta x|)\\
		\widehat {\Delta y} = sign(y) \cdot \log (1 + |\Delta y|)
	\end{array}
\end{equation}
\begin{equation}
	\vspace{-0.1cm}
	\label{equation16}
	B(\Delta x,\Delta y) = \Phi (\Delta x,\Delta y)
\end{equation}
${\Phi}$ is the 2-layer MLP with a ReLU activation. ${\Delta x}$, ${\Delta y}$ and ${\widehat {\Delta x}}$, ${\widehat {\Delta y}}$ are the linear-scaled and log-spaced coordinates, respectively.
\vspace{-1pt}
\subsection{Mixed Feature Pyramid}
\label{section3_3}
\begin{figure*}[t]
	\setlength{\belowcaptionskip}{-0.1cm}
	\centering
	\includegraphics[width=0.8\linewidth]{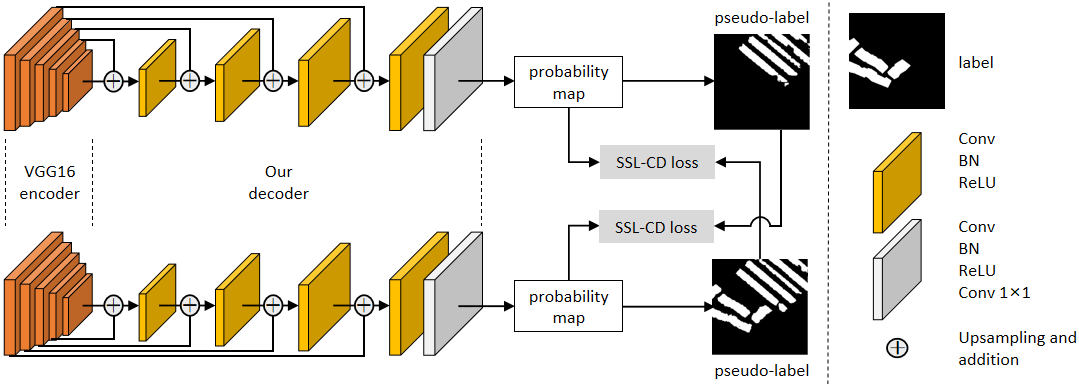}
	\caption{CNN branch network and SSL strategy.}
	\label{fig5}
\end{figure*}
FPT utilizes different transformers to provide the self-attention features of different layers and the interaction features between layers \cite{Zhang2020FeaturePT}. Self-transformer provides the self-attention features of three layers. Rendering transformer provides a down-top interaction feature between layers, and grounding transformer provides a top-down interaction feature. Inspired by FPT and the requirements of the existing task, we construct a novel multi-scale module, MFP, by improving FPT. 

Since the basic blocks of current networks have evolved from convolutions to transformers, we remove self-transformer. Moreover, we delete rendering transformer due to it does not experimentally play a role on the VHRRS images change detection. Thus we only impose grounding transformer to provide the interaction information between layers. However, in addition to multi-scale features of different layers and multi-scale interaction features between layers, multi-scale information also includes multi-scale features modeling within layers. So we use SK-Conv \cite{li2019selective} and Res2Net-Conv \cite{gao2019res2net} to provide the multi-scale features inside different layers, respectively. Of course, other multi-scale convolutions, for example, inception, dilated convolution, can be used here. We only use SK-Conv and Res2Net-Conv as a example to experimentally validate our idea. MFP, FPT, SK-Conv and Res2Net-Conv are shown in Fig. \ref{fig4}.

Firstly, SK-Conv and Res2Net-Conv are severally used to mine the intra-layer multi-scale information of the input features ${a_f}$, ${b_f}$, and ${c_f}$. At the same time, grounding transformer is utilized to provide the multi-scale interaction features between different layers. These features are then rearranged according to different spatial resolutions. After residual blocks are added to the different layers, feature fusion is performed. Finally, using convolution, CBAM, and dropout in turn, we get the output features with intra-layer multi-scale information and multi-scale interaction information between layers. The roles of CBAM and dropout are to enhance the effective fusion of different features and prevent overfitting, respectively. Since the spatial and spectral resolution of the input features are consistent with the output, MFP is a plug-and-play module. MFP has also been shown to be effective for a variety of VHRRS images change detection models.

Finally, we introduce the three multi-scale feature extractors used. The core operation of grounding transformer (GT) is:
\begin{equation}
	\vspace{-0.1cm}
	\label{equation17}
	GT({Q_g},{K_g},{V_g}) = Softmax({Q_g}K_g^T){V_g}
\end{equation}
Here, ${Q_g}$, ${K_g}$ and ${V_g}$ are the nonlinear transformations of ${X_g}$. ${K_g^T}$ is the transposition matrix of ${K_g}$. And we define that ${X_g}$, ${\bar X_g}$ and ${\tilde X_g}$ are the input feature, intermediate variable and output feature of GT, respectively. Take the layer ${a_m}$ and ${b_m}$ in Fig. \ref{fig4} as an example to explain the complete operation of GT:
\begin{equation}
	\vspace{-0.1cm}
	\label{equation18}
		\begin{array}{l}
			\bar X_g^a = Up(BN(Conv(X_g^a)))\\
			\bar X_g^b = BN(Conv(X_g^b))\\
			\tilde X_g^{ab} = GT(Cat(\bar X_g^a,\bar X_g^b))
		\end{array}
\end{equation}
Both SK-Conv and Res2Net-Conv are the multi-scale convolutions. SK-Conv obtains the features ${U_s}$ and ${V_s}$ by a 3${\times}$3 convolution and a 5${\times}$5 convolution, respectively. A joint SE attention is then applied for ${U_s}$ and ${V_s}$ to obtain ${\bar U_s}$ and ${\bar V_s}$. In the end, ${\bar U_s}$ plus ${\bar V_s}$ gives the SK-Conv feature ${\tilde X_s}$. The multi-scale component of Res2Net-Conv can be expressed:
\begin{equation}
	\vspace{-0.1cm}
	\label{equation19}
		y_r^z = \left\{ \begin{array}{l}
			x_r^z{\rm{\quad \quad \quad \quad \quad \quad \quad \quad z = 1}}\\
			Conv(x_r^z){\rm{\quad \quad \quad \quad \quad z = 2}}\\
			Conv(x_r^z + y_r^{z - 1}){\rm{\quad 2 < z}} \le {\rm{4}}
		\end{array} \right.
\end{equation}
We pass the input ${X_r}$ through 1${\times}$1 convolution, Eq. (\ref{equation19}), and 1${\times}$1 convolution in turn, to get the Res2Net-Conv feature ${\tilde X_r}$.

\subsection{CNN Branch Network and SSL Strategy}
\label{section3_4}
To solve the problem that the VGG16 encoder \cite{simonyan2014very} is untrainable, we design a new decoder and impose the SSL strategy \cite{chen2022fccdn} to optimize the low-level features of the VGG16 encoder. The detailed implementation is shown in Fig. \ref{fig5}. We compose the decoder base block with convolution, BN and ReLU. The five base blocks are arranged in order of the increasing spatial resolutions. Further more, the decoder features and the corresponding encoder features are fused using upsampling and addition. Finally, 1${\times}$1 convolution is used to output the probability map of change detection.

For SSL strategy, we firstly generate the pseudo-labels based on the probability maps through CNN branch network:
\begin{equation}
	\vspace{-0.1cm}
	\label{equation20}
	\begin{array}{l}
		P{L_{1,{\rm{u}}}} = \left\{ \begin{array}{l}
			0\quad P{M_{1,{\rm{u}}}} < 0.5\\
			1\quad P{M_{1,{\rm{u}}}} \ge 0.5
		\end{array} \right.\\
		P{L_{2,{\rm{u}}}} = \left\{ \begin{array}{l}
			0\quad P{M_{2,{\rm{u}}}} < 0.5\\
			1\quad P{M_{2,{\rm{u}}}} \ge 0.5
		\end{array} \right.
	\end{array}
\end{equation}
Here, ${u}$ refers to each pixel of the probability maps or pseudo-labels. ${P{M_{1,{\rm{u}}}}}$ and ${P{M_{2,{\rm{u}}}}}$ are the two probability maps that are the outputs from the branch network. Since ${P{M_{1,{\rm{u}}}}}$ and ${P{M_{2,{\rm{u}}}}}$ are normalized to [0,1] by sigmoid, we can obtain the pseudo-labels ${P{L_{1,{\rm{u}}}}}$ and ${P{L_{2,{\rm{u}}}}}$ via Eq. (\ref{equation20}).
After that, we determine the changed and unchanged regions based on the labels from the change detection datasets. In the unchanged regions, we use the pseudo-labels generated by the one branch to supervise the other branch. In the changed regions, we use the opposite results of the pseudo-labels from the one branch, to supervise the other branch. The goal is to keep the unchanged features as close as possible and the changed features as far away as possible. Therefore, the encoder features provided by CNN branch network are rich in semantic information and have more discrimination ability for change detection. The SSL-CD loss is designed as follows:
\begin{equation}
	\vspace{-0.1cm}
	\label{equation21}
	\begin{array}{l}
		{L_{SSL1}} = F(P{M_{1,{\rm{u}}}},P{L_{2,{\rm{u}}}}\left| {u \in {U_{\alpha }}} \right.)\\\quad \quad \quad \quad \quad  + F(P{M_{1,{\rm{u}}}},1 - P{L_{2,{\rm{u}}}}\left| {u \in {C_{\alpha }}} \right.)\\
		{L_{SSL2}} = F(P{M_{2,{\rm{u}}}},P{L_{1,{\rm{u}}}}\left| {u \in {U_{\alpha }}} \right.)\\\quad \quad \quad \quad \quad  + F(P{M_{2,{\rm{u}}}},1 - P{L_{1,{\rm{u}}}}\left| {u \in {C_{\alpha }}} \right.)
	\end{array}
\end{equation}
where ${{U_{\alpha }}}$ and ${{C_{\alpha }}}$ represent respectively the unchanged and changed regions. ${{L_{SSL1}}}$ and ${{L_{SSL2}}}$ are the two SSL-CD loss. ${F()}$ is a metric function, and we choose ${{L_{BCE}}}$ as it.

\subsection{Loss Function}
\label{section3_5}
First, bi-temporal change detection is fundamentally a binary classification task, so binary cross entropy (BCE) loss is usually used:
\begin{equation}
	\vspace{-0.1cm}
	\label{equation22}
	{L_{BCE}} =  - (t\log (\hat t) + (1 - t)\log (1 - \hat t))
\end{equation}
In our paper, ${t}$ and ${\hat t}$ denote the predicted change confidence and the label in the corresponding position, respectively. For change detection tasks, however, the changed regions are far less than the unchanged regions, so there is a serious class imbalance problem in change detection. For example, the ratio of changed pixels to unchanged pixels in CDD is 0.147 \cite{qin2013object}. To mitigate this problem, Dice loss is often used:
\begin{equation}
	\vspace{-0.1cm}
	\label{equation23}
	{L_{Dice}} = 1 - \frac{{2\hat tt + \sigma }}{{\hat t + t + \sigma }}
\end{equation}
Here, adding ${\sigma}$ avoids the case where the denominator is zero.

The loss function used by our model is a combination of BCE and Dice loss. Moreover, to address the features shift during the training of the large network, we use the deep supervision strategy. Specifically, the deep supervision strategy uses the same labels as those used for the network output. The labels are replicated in four copies for the four deep supervision interfaces. And DConvs ${C_{0,1}}$, ${C_{0,2}}$, ${C_{0,3}}$, ${C_{0,4}}$ output the probability maps through convolution and sigmoid function. ${\sum\limits_{m = 1}^4 {L_{BCE}^m}  + {\lambda _1}L_{Dice}^m}$, at last, is used to measure and optimize these probability maps by the deep supervision labels. The total loss function is expressed as follows:
\begin{equation}
	\vspace{-0.1cm}
	\label{equation24}
	{L_{Total}} = \sum\limits_{m = 1}^5 {L_{BCE}^m}  + {\lambda _1}L_{Dice}^m + {\lambda _2}{L_{SSL1}} + {\lambda _2}{L_{SSL2}}
\end{equation}
where ${\sum\limits_{m = 1}^5  {L_{BCE}^m}  + {\lambda _1}L_{Dice}^m}$ represents the output loss and the four deep supervision losses of main network. ${L_{SSL1}}$ and ${L_{SSL2}}$ are the self-supervision losses of CNN branch network. ${\lambda _1}$ and ${\lambda _2}$ are the weight coefficients. ${\lambda _1}$ is set to 0.5, and ${\lambda _2}$ is 0.25.

\section{Experimental results and analysis}
\label{section4}
\subsection{Experimental Configurations}
\label{section4_1}
\subsubsection{Datasets}
our model is tested on the four publicly available change detection datasets, achieving the SOTA results. Due to GPU memory limitations, we crop the images into the non-overlapping patchs of size 256×256 for the four datasets.
\begin{itemize}
	\item[(1)] Learning, vision, and remote sensing change detection dataset (LEVIR-CD) \cite{chen2020spatial} is a publicly available change detection resource for large buildings. It comprises 637 pairs of high-resolution (0.5 m) remote sensing images, each measuring 1024×1024 pixels. We routinely crop these images and then obtain 7120/1024/2048 pairs as the training/validation/test data.
	\item[(2)] Change detection dataset (CDD) \cite{lebedev2018change} contains 11 pairs of multi-spectral images obtained from Google Earth with spatial resolutions ranging from 0.03 to 1 m. Following the dataset partitioning, we apply the data augmentation methods, image rotation and image flipping, to the training set. As a result, we obtain a total of 60000/3000/3000 training/validation/test pairs.
	\item[(3)] Wuhan university change detection dataset (WHU-CD) \cite{ji2018fully} focuses on the buildings change detection. It features a pair of high-resolution (0.075 m) aerial images, each measuring 32507×15354 pixels. Since there is not a general data partitioning scheme for WHU-CD, we cut these images into the non-overlapping segments of size 256×256 and randomly divide them into 6096/764/764 pairs for the training/validation/testing sessions, respectively.
	\item[(4)] Sun Yat-sen university change detection dataset (SYSU-CD) \cite{shi2021deeply} contains 20000 pairs of orthographic aerial images with the spatial resolution of 0.5 m taken in Hong Kong. Each image is 256×256 pixels. We use the 12000/4000/4000 training/validation/testing pairs based on the dataset provider splitting. It is worth noting that SYSU-CD presents multiple types of changed objects in the more complex scenario, making it a particularly challenging dataset.
\end{itemize}

\subsubsection{Baseline and State-of-the-art Methods}
we compare SwinV2DNet with the baseline and state-of-the-art methods as follows. The first three serve as the baselines, while the last seven represent the advanced networks developed over the past three years. We implement these change detection networks using the publicly available codes and default hyperparameters.
\begin{table*}[h!]
	\caption{\centering{The comparison results on the three change detection datasets. The best values are\newline highlighted in bold font. All results are expressed as percentages ($\%$).}}
	
	\centering
	\scalebox{0.85}{
		\renewcommand\arraystretch{1}
		\setlength{\tabcolsep}{3mm}{
			\begin{tabular}{c|ccc}
				\hline
				\multirow{2}{*}{\textbf{Method}} & \textbf{LEVIR-CD}                     & \textbf{CDD}                & \textbf{WHU-CD}             \\
				& Pre. / Rec. / F1 / IoU / OA    & Pre. / Rec. / F1 / IoU / OA    & Pre. / Rec. / F1 / IoU / OA \\ \hline
				FC-EF                            & 86.16 / 86.20 / 86.18 / 76.16 / 98.59 & 85.35 / 77.56 / 81.27 / 42.14 / 95.59     & 86.13 / 86.01 / 86.07 / 75.67 / 98.82 \\
				FC-Siam-Diff                     & 90.36 / 84.81 / 87.50 / 81.06 / 98.77 & 92.28 / 78.70 / 84.95 / 48.87 / 96.56     & 81.40 / 89.11 / 85.08 / 71.79 / 98.68 \\
				FC-Siam-Conc                     & 87.30 / 87.81 / 87.55 / 75.09 / 98.73 & 92.04 / 81.94 / 86.70 / 52.95 / 96.90     & 79.98 / 90.94 / 85.11 / 66.25 / 98.65 \\ \hline
				IFNet                            & \textbf{93.73} / 87.31 / 90.40 / 84.04 / 99.06 & 97.71 / 93.64 / 95.63 / 81.31 / 98.94     & \textbf{98.51} / 82.46 / 89.77 / 90.28 / 99.21 \\
				SNUNet-CD                        & 91.00 / 88.30 / 89.63 / 79.18 / 98.96 & 98.13 / 97.62 / 97.87 / 88.24 / 99.48     & 88.50 / 90.31 / 89.40 / 73.50 / 99.09 \\
				BIT                              & 91.81 / 88.00 / 89.86 / 79.35 / 98.99 & 97.07 / 96.43 / 96.75 / 83.52 / 99.20     & 92.10 / 92.41 / 92.26 / 78.56 / 99.34 \\
				DCFF-Net                         & 92.96 / 89.83 / 91.37 / 82.80 / 99.14 & 98.75 / 98.94 / 98.84 / 92.79 / 99.71     & 96.51 / 93.11 / 94.78 / 88.55 / 99.57 \\
				TransUNetCD                      & 90.62 / 88.44 / 89.52 / 79.63 / 98.94 & 97.44 / 96.52 / 96.98 / 84.81 / 99.26     & 94.72 / 91.21 / 92.93 / 83.82 / 99.41 \\
				ICIF-Net                         & 92.23 / 88.53 / 90.34 / 81.54 / 99.04 & 97.61 / 97.04 / 97.32 / 85.98 / 99.34     & 93.61 / 89.69 / 91.61 / 82.61 / 99.31 \\
				FCCDN                            & 92.10 / 84.86 / 88.33 / 80.48 / 98.86 & 95.96 / 95.56 / 95.76 / 79.79 / 98.96     & 92.65 / 90.44 / 91.53 / 79.94 / 99.29 \\ \hline
				Ours                             & 92.98 / \textbf{91.33} / \textbf{92.15} / \textbf{85.44} / \textbf{99.21} & \textbf{99.18} / \textbf{99.12} / \textbf{99.15} / \textbf{94.02} / \textbf{99.79}     & 96.75 / \textbf{94.65} / \textbf{95.69} / \textbf{90.34} / \textbf{99.64} \\ \hline
		\end{tabular}}
	}
	\label{tab1}
\end{table*}
\begin{itemize}
	\item[(1)] FC-EF \cite{daudt2018fully}: bi-temporal change detection images are concatenated as a single input to FCN.
	\item[(2)] FC-Siam-Diff \cite{daudt2018fully}: a siamese FCN is employed to extract the multi-level features, utilizing the differences of these features to detect the changed information.
	\item[(3)] FC-Siam-Conc \cite{daudt2018fully}: the multi-level features are extracted and fused using a siamese FCN with the cascaded architecture.
	\item[(4)] IFNet \cite{zhang2020deeply}: CBAM is applied to the heterogeneous features at each level of the cascaded decoder, and deep supervision is used for the improved training of intermediate layers.
	\item[(5)] SNUNet-CD \cite{fang2021snunet}: a combination of siamese structure and UNet++ is utilized to extract the high-level features.
	\item[(6)] BIT \cite{chen2021remote}: a serial hybrid network that embeds transformer into ResNet.
	\item[(7)] DCFF-Net \cite{pan2021dcff}: a parallel pure CNN that combines VGG16 with UNet++, and integrates CBAM and deep supervision.
	\item[(8)] TransUNetCD \cite{li2022transunetcd}: a serial cascaded hybrid network that embeds transformer into UNet.
	\item[(9)] ICIF-Net \cite{feng2022icif}: a parallel hybrid network focusing on the interaction and fusion of transformer and CNN.
	\item[(10)] FCCDN \cite{chen2022fccdn}: the supervised model with self-supervised strategy, producing the features rich in semantic information.
\end{itemize}

\subsubsection{Implementation Details}
we implement our model using PyTorch and train it on a single NVIDIA GeForce RTX 3090 GPU. During the training, we optimize the model with Adam optimizer. The batch size is set to 4. The learning rate is initially set to ${5\times {10^{-5}}}$ and linearly decays to 0 over the course of 200 epochs. 

\subsubsection{Evaluation Metrics}
F1-score is an index used to measure the performance of binary classification models, taking into account both precision and recall. So we primarily employ F1-score with respect to the change category as the main evaluation metric. F1-score is shown below:
\begin{equation}
	F1 = \frac{{2 \times TP}}{{2 \times TP + FP + FN}}
\end{equation}
In addition, we also report precision, recall, intersection over union (IoU) for the change category, and overall accuracy (OA). These metrics are defined as follows:
\begin{equation}
	\vspace{-0.3cm}
    {\rm{precision = }}\frac{{TP}}{{TP + FP}}
\end{equation}
\begin{equation}
	\vspace{-0.3cm}
	{\rm{recall = }}\frac{{TP}}{{TP + FN}}
\end{equation}
\begin{equation}
	\vspace{-0.3cm}
	{\rm{IoU = }}\frac{{TP}}{{TP + FP + FN}}
\end{equation}
\begin{equation}
	{\rm{OA = }}\frac{{TP + TN}}{{TP + TN + FP + FN}}
\end{equation}
Here, TP, TN, FP, and FN represent the number of true positive, true negative, false positive, and false negative, respectively.

\subsection{Experimental Results}
\subsubsection{Results Analysis and Comparison}
\begin{table}[h!]
	\caption{\centering{The comparison results on SYSU-CD dataset. The best values are highlighted in bold font. All results \newline are expressed as percentages ($\%$).}}
	\centering
	\scalebox{0.85}{
		\renewcommand\arraystretch{1}
		\setlength{\tabcolsep}{3mm}{
			\begin{tabular}{c|c}
				\hline
				\multirow{2}{*}{\textbf{Method}} 
				& \textbf{SYSU-CD}      \\
				& Pre. / Rec. / F1 / IoU / OA     \\ \hline
				FC-EF                  & 79.30 / 68.84 / 73.70 / 44.64 / 88.41 \\
				FC-Siam-Diff           & \textbf{89.80} / 58.49 / 70.84 / 42.37 / 88.64 \\
				FC-Siam-Conc           & 82.31 / 73.52 / 77.67 / 50.33 / 90.03 \\ \hline
				IFNet                  & 85.16 / 75.36 / 79.96 / 57.22 / 91.09 \\
				SNUNet-CD              & 80.03 / 76.62 / 78.29 / 52.99 / 89.98 \\
				BIT                    & 81.67 / 76.52 / 79.01 / 52.41 / 90.41 \\
				DCFF-Net               & 78.71 / \textbf{86.24} / 82.30 / 62.05 / 91.25 \\
				TransUNetCD            & 77.25 / 80.17 / 78.68 / 54.72 / 89.75 \\
				ICIF-Net               & 78.53 / 78.89 / 78.71 / 53.80 / 89.94 \\
				FCCDN                  & 78.57 / 78.14 / 78.36 / 53.33 / 89.82 \\ \hline
				Ours                   & 83.46 / 82.81 / \textbf{83.13} / \textbf{62.90} / \textbf{92.08} \\ \hline
		\end{tabular}}
	}
	\label{tab2}
\end{table}
Tables \ref{tab1} and \ref{tab2} present the overall comparison results for the test sets: LEVIR-CD, CDD, WHU-CD, and SYSU-CD. Through quantitative analysis, our model has demonstrated the significant improvements over other methods in the three key indicators, F1 score, IoU, and OA, across these datasets. Notably, our method outperforms the recent DCFF-Net by 0.78/0.31/0.91/0.83 in F1 score, underscoring the significance of both global information provided by transformer and local information represented by CNN for change detection. Furthermore, our approach also showcases a performance advantage over the two serial networks, BIT and TransUNetCD, reinforcing the superiority of the parallel combination of transformer and CNN. In summary, our proposed model achieves the SOTA results by leveraging a parallel architecture of transformer and CNN with the multi-scale and self-supervision features that enhance its discrimination capabilities (refer to the analysis about Table \ref{tab3}).
\begin{figure*}[h]
	\centering
	\subfloat{
		\includegraphics[width=0.99\linewidth]{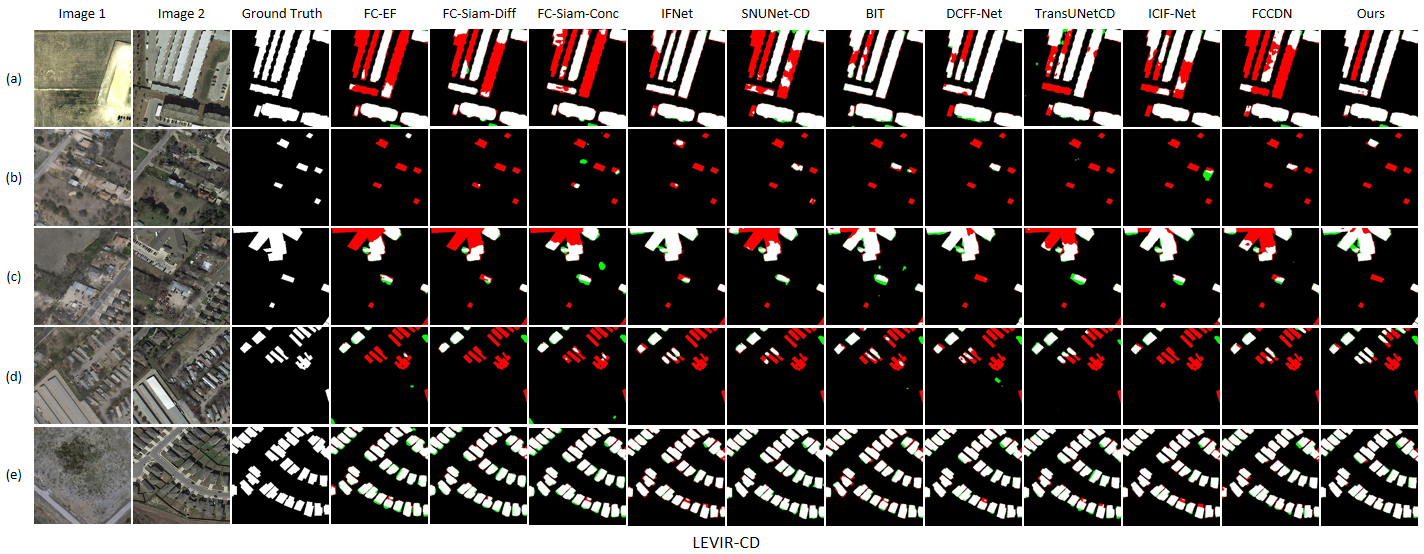}}
	
	\vspace{-0.01cm}
	\subfloat{
		\includegraphics[width=0.99\linewidth]{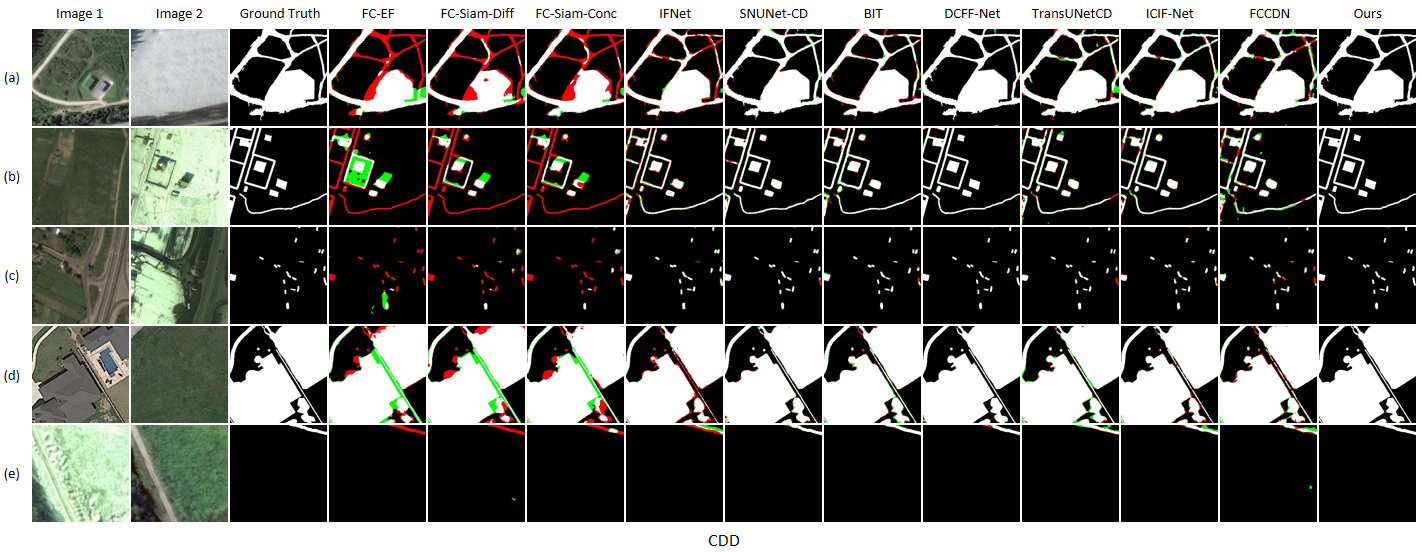}}
	
	\caption{The visualization results of various methods on LEVIR-CD and CDD test sets. We use different colors to represent TP(white), TN(black), FP(green), and FN(red) in the change maps. (a)${\sim}$(e) show the prediction results of these methods for different samples, respectively.}\label{fig6}
\end{figure*}
\begin{figure*}[h]
	\centering
	\subfloat{
		\includegraphics[width=0.99\linewidth]{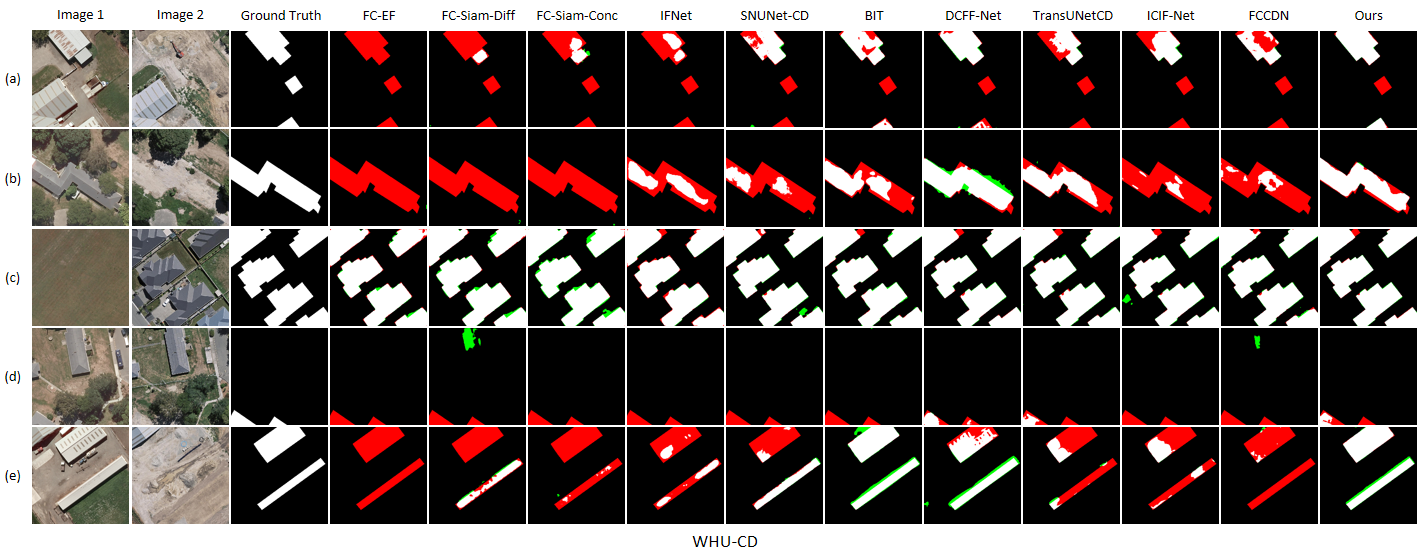}}
	
	\vspace{-0.01cm}
	\subfloat{
		\includegraphics[width=0.99\linewidth]{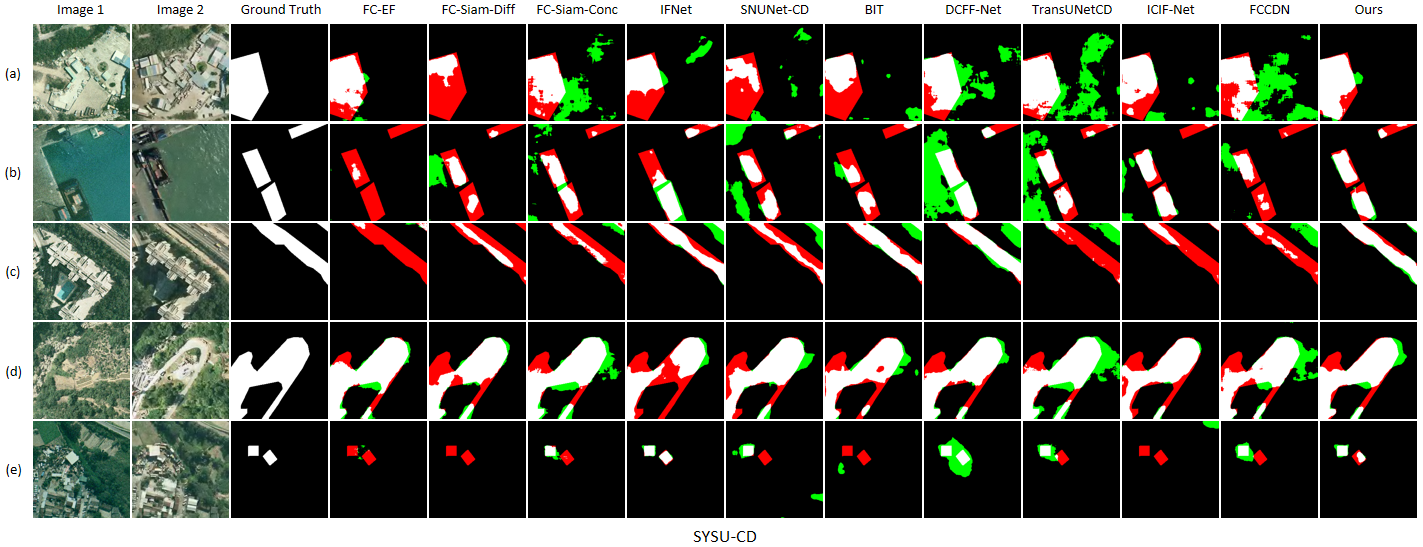}}
	
	\caption{The visualization results of various methods on WHU-CD and SYSU-CD test sets. We use different colors to represent TP(white), TN(black), FP(green), and FN(red) in the change maps. (a)${\sim}$(e) show the prediction results of these methods for different samples, respectively.}\label{fig7}
\end{figure*}

The comparison of visualization results on the four datasets is shown in Figs. \ref{fig6} and \ref{fig7}. We use different colors to represent TP(white), TN(black), FP(green), and FN(red) in the change maps. It is observed that our model maintains the best structured state compared to other models from (a), (c) of LEVIR-CD and (a), (b), (e) of WHU-CD. Additionally, our model demonstrates the superior performance in detecting dense small objects and edge objects, as evidenced by LEVIR-CD (d) and WHU-CD (d). 
The CDD dataset poses the unique challenges due to its varying illumination conditions and high occurrence of pseudo changes, such as the snow cover is added in the image 2 of (a), and the illumination differences are visible in the (b)${\sim}$(e) image pairs. However, our model outperforms other methods in these difficult scenes, particularly for the three detection requirements of structured changed regions, small targets, and edge targets.
The challenge of SYSU-CD dataset lies also in the high intraclass variation and low interclass variance of background and targets. Our model is still able to maintain the structure of targets relatively well in the scenes where the targets are cluttered and close to the background.
The detection result are well understood. Most of the structured regions are obtained from the global information of transformer, and the description of small targets and edge targets comes from the low-level detail information of CNN. This once again validates the importance of the parallel dense architecture of transformer and CNN.

\subsubsection{Training Processes Analysis}
\begin{figure*}[t]
	\vspace{-0.4cm}
	\centering
	\includegraphics[width=0.85\linewidth]{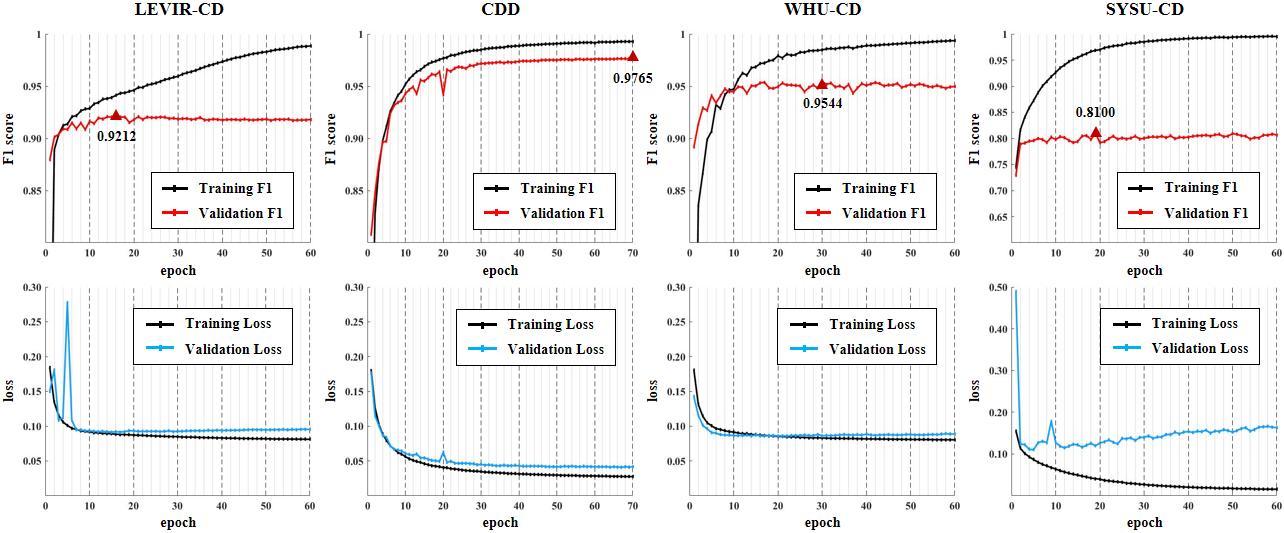}
	\caption{The training processes analysis of SwinV2DNet.}
	\label{fig8}
\end{figure*}

we evaluate the performance of SwinV2DNet on the four datasets by tracking the two metrics, F1 score and loss, as depicted in Fig. \ref{fig8}. The F1 score and loss curves provide an intuitive understanding that our model is stably convergent and elegantly efficient. The peak values on the F1 curves are observed at points 0.9212(LEVIR-CD), 0.9765(CDD), 0.9544(WHU-CD), and 0.8100(SYSU-CD), indicating that our model requires a training process of just 35 epochs. Given the CDD dataset's abundance of small targets and intricate labelings, SwinV2DNet exhibits a slight growth even after 35 epochs. Comparing the datasets from a horizontal perspective, SYSU-CD appears to be the most challenging and prone to the rapid overfitting. This could potentially be attributed to the significantly differing distribution between training data and validation data, wherein the overfitting on training data impairs the model's generalization capacity. Currently, no relevant studies have been conducted to explain this phenomena.
\begin{table*}[h!]
	\caption{\centering{The ablation studies for the overall network on the three datasets. We report F1 and OA scores. The base and best results are annotated in blue and red, respectively. All results are expressed as percentages ($\%$).}}
	\centering
	\scalebox{0.85}{
		\renewcommand\arraystretch{1}
		\begin{tabular}{cccccc}
			\hline
			\multicolumn{3}{c}{\makebox[0.24\textwidth][c]{\textbf{Overall Network}}}                  
			& \textbf{LEVIR-CD} 
			& \textbf{CDD}                    & \textbf{WHU-CD}   \\ 		 
			\makebox[0.09\textwidth][c]{Swin-V2}          & \makebox[0.01\textwidth][c]{MFP}                
			& SSL            
			& F1 / OA                                    
			& F1 / OA                              
			& F1 / OA   \\ \hline
			&                    &             
			& \bf{\textcolor[rgb]{0 0.44 0.75}{91.37}} / \bf{\textcolor[rgb]{0 0.44 0.75}{99.14}}      
			& \bf{\textcolor[rgb]{0 0.44 0.75}{94.81}} / \bf{\textcolor[rgb]{0 0.44 0.75}{98.73}}      
			& \bf{\textcolor[rgb]{0 0.44 0.75}{94.78}} / \bf{\textcolor[rgb]{0 0.44 0.75}{99.57}}   \\ \hline
			\checkmark           & \multicolumn{1}{c}{}  & \multicolumn{1}{c}{} 
			& 91.85 / 99.18   
			& 97.58 / 99.41     
			& 95.34 / 99.61   \\	
			\multicolumn{1}{c}{} & \checkmark            & \multicolumn{1}{c}{}
			& 91.47 / 99.15   
			& 96.15 / 99.06          
			& 94.80 / 99.57   \\	
			\multicolumn{1}{c}{} & \multicolumn{1}{c}{}  & \checkmark                    
			& 91.38 / 99.13    
			& 96.66 / 99.18     
			& 95.25 / 99.60   \\ \hline
			\checkmark           & \checkmark            & \multicolumn{1}{c}{} 
			& 91.97 / 99.19
			& 97.66 / 99.43
			& 95.10 / 99.59   \\
			\checkmark           & \multicolumn{1}{c}{}  & \checkmark                    
			& 92.00 / 99.19
			& 97.60 / 99.41
			& 95.31 / 99.61   \\
			\multicolumn{1}{c}{} & \checkmark            & \checkmark                   
			& 91.40 / 99.13
			& 96.69 / 99.19
			& 94.82 / 99.57   \\ \hline			
			\checkmark           & \checkmark            & \checkmark                    
			& \bf{\textcolor[rgb]{1 0 0}{92.15}} / \bf{\textcolor[rgb]{1 0 0}{99.21}}
			& \bf{\textcolor[rgb]{1 0 0}{97.71}} / \bf{\textcolor[rgb]{1 0 0}{99.44}}
			& \bf{\textcolor[rgb]{1 0 0}{95.69}} / \bf{\textcolor[rgb]{1 0 0}{99.64}}   \\ \hline
		\end{tabular}
	}
	\label{tab3}
\end{table*}

The further conjoint analysis with the testing F1 values of 0.9215(LEVIR-CD), 0.9771(CDD), 0.9569(WHU-CD), and 0.8313(SYSU-CD) in Tables \ref{tab1} and \ref{tab2} reaffirms the strong generalization ability of our model. 0.9771(CDD) reveals the F1 performance for the CDD test set without applying any data augmentations. Based on the compounded features, dense connection and deep supervision in our model greatly contribute to the stability and efficiency of training processes.
\vspace{-0.4cm}
\subsection{Ablation Studies and  Parameter Analysis}
\subsubsection{Ablation Study of Overall Network}
\begin{table*}[h!]
	\caption{\centering{The longitudinal dissection (left) and horizontal promotion (right) for MFP on the LEVIR-CD dataset. The base and best results are annotated in blue and red, respectively. All results are expressed as percentages ($\%$).}}
	\begin{minipage}[c]{0.5\textwidth}
		\centering
		\scalebox{0.85}{
			\renewcommand\arraystretch{1}
			\begin{tabular}{ccccc}
				\hline
				\multicolumn{3}{c}{\textbf{MFP}}                    
				& \multicolumn{1}{c}{\textbf{FC-Siam-Diff}}            
				& \multicolumn{1}{c}{\textbf{Ours}}   \\ 		 
				GT            & Res2Net-Conv           & SK-Conv                                                 
				& F1 / OA                              
				& F1 / OA   \\ \hline
				&                          &                 
				& \bf{\textcolor[rgb]{0 0.44 0.75}{87.50}} / \bf{\textcolor[rgb]{0 0.44 0.75}{98.77}}          
				& \bf{\textcolor[rgb]{0 0.44 0.75}{91.845}} / \bf{\textcolor[rgb]{0 0.44 0.75}{99.177}}   \\ \hline			
				\checkmark        & \multicolumn{1}{c}{}     & \multicolumn{1}{c}{}                                
				& 87.49 / 98.74                         
				& 91.955 / 99.192      \\			
				\multicolumn{1}{c}{} & \checkmark            & \multicolumn{1}{c}{}                             
				& 87.89 / 98.79                        
				& 92.057 / 99.191      \\			
				\multicolumn{1}{c}{} & \multicolumn{1}{c}{}  & \checkmark                                             
				& 87.66 / 98.74                         
				& 91.915 / 99.186      \\ \hline			
				\checkmark        & \checkmark               & \multicolumn{1}{c}{}                              
				& 87.78 / 98.78    
				& 91.975 / 99.190      \\			
				\checkmark        & \multicolumn{1}{c}{}     & \checkmark                    
				& 87.50 / 98.75
				& 91.866 / 99.178   \\			
				\multicolumn{1}{c}{} & \checkmark            & \checkmark                   
				& 87.92 / 98.78
				& 91.843 / 99.179   \\ \hline			
				\checkmark        & \checkmark               & \checkmark                    
				& \bf{\textcolor[rgb]{1 0 0}{88.20}} / \bf{\textcolor[rgb]{1 0 0}{98.81}}
				& \bf{\textcolor[rgb]{1 0 0}{92.060}} / \bf{\textcolor[rgb]{1 0 0}{99.200}}   \\ \hline			
			\end{tabular}
		}
	\end{minipage}
	\begin{minipage}[c]{0.5\textwidth}
		\centering
		\scalebox{0.85}{
			\renewcommand\arraystretch{1}
			\setlength{\tabcolsep}{3mm}{
				\begin{tabular}{ccc}
					\hline
					\multirow{2}{*}{\textbf{Method}}
					& \multirow{2}{*}{\textbf{MFP}} 
					& \textbf{LEVIR-CD}     \\ 
					&& Pre. / Rec. / F1 / IoU / OA  \\ \hline		    
					FC-Siam-Conc  &                      
					& 87.30 / 87.81 / \bf{\textcolor[rgb]{0 0.44 0.75}{87.55}} / \bf{\textcolor[rgb]{0 0.44 0.75}{75.09}} / \bf{\textcolor[rgb]{0 0.44 0.75}{98.73}} \\
					FC-Siam-Conc  & \checkmark           
					& 87.91 / 88.73 / \bf{\textcolor[rgb]{1 0 0}{88.32}} / \bf{\textcolor[rgb]{1 0 0}{77.04}} / \bf{\textcolor[rgb]{1 0 0}{98.80}} \\ \hline
					IFNet         &   				                   
					& 93.73 / 87.31 / \bf{\textcolor[rgb]{0 0.44 0.75}{90.40}} / \bf{\textcolor[rgb]{0 0.44 0.75}{84.04}} / \bf{\textcolor[rgb]{0 0.44 0.75}{99.06}} \\
					IFNet         & \checkmark           
					& 93.68 / 88.02 / \bf{\textcolor[rgb]{1 0 0}{90.76}} / \bf{\textcolor[rgb]{1 0 0}{84.44}} / \bf{\textcolor[rgb]{1 0 0}{99.09}} \\ \hline
					SNUNet-CD           &                      
					& 91.00 / 88.30 / \bf{\textcolor[rgb]{0 0.44 0.75}{89.63}} / \bf{\textcolor[rgb]{0 0.44 0.75}{79.18}} / \bf{\textcolor[rgb]{0 0.44 0.75}{98.96}} \\
					SNUNet-CD           & \checkmark           
					& 90.35 / 89.07 / \bf{\textcolor[rgb]{1 0 0}{89.70}} / \bf{76.64} / \bf{\textcolor[rgb]{1 0 0}{98.96}} \\ \hline
					FCCDN         &                      
					& 92.10 / 84.86 / \bf{\textcolor[rgb]{0 0.44 0.75}{88.33}} / \bf{\textcolor[rgb]{0 0.44 0.75}{80.48}} / \bf{\textcolor[rgb]{0 0.44 0.75}{98.86}} \\ 
					FCCDN         & \checkmark           
					& 89.95 / 87.66 / \bf{\textcolor[rgb]{1 0 0}{88.79}} / 79.91 / \bf{\textcolor[rgb]{1 0 0}{98.87}} \\ \hline
			\end{tabular}}
		}
	\end{minipage}
	\label{tab5}
\end{table*}
in the overall network architecture, our contributions consist of the three parts: Swin V2 main network, MFP, and CNN branch network trained using SSL strategy. As shown in Table \ref{tab3}, we perform the ablation studies for these three contributions on the LEVIR-CD, CDD, and WHU-CD datasets. According to the data analysis of Table \ref{tab3}, Swin V2 main network has the largest effect on the overall performance improvement, and the latter two have the similar effects. Especially on the CDD dataset, Swin V2 main network shows a improvement of F1 score 2.77$\%$ compared to CNN main network. And the different combinations of two contributions occasionally occur the mutually exclusive phenomenas. However, our model achieves the significantly better results than the baseline on the three datasets, using the three contributions together. Our ablation experiments use F1 score as the main evaluation metric that is roughly positively correlated with OA. This also verifies the importance of our parallel compounded architecture consisting of Swin V2 main network and CNN branch network. The three types of multi-scale features and the encoder semantic features guided by SSL strategy also have clear guiding effects on the model performance.

\subsubsection{Parameter Analysis of Swin V2}
for Swin V2 blocks, we mainly perform the ablation studies about the pretrained weights and number of blocks as described in Table \ref{tab4}. We only use two Swin V2 blocks for ${S_{1,1}}$, ${S_{1,2}}$, and ${S_{2,1}}$. For the U-shaped structure formed by [${S_{1,0}}$, ${S_{2,0}}$, ${S_{3,0}}$, ${S_{4,0}}$, ${S_{3,1}}$, ${S_{2,2}}$, ${S_{1,3}}$], we try the three configurations of Swin V2 blocks: [2, 2, 2, 2, 2, 2, 2], [2, 2, 6, 2, 6, 2, 2], and [2, 2, 18, 2, 18, 2, 2]. The ablation results on the three datasets show that the combination of the pre-trained weights and configuration of [2, 2, 6, 2, 6, 2, 2] has the advanced detection performance and robust application scenarios. This is due to the facts that the pretrained weights usually contain the prior information of upstream tasks, and too many Swin V2 blocks maybe lead to the underfitting of some intermediate layer parameters of the model.
\begin{table}[h!]
	\caption{\centering{The ablation studies for Swin V2 on the three datasets. We report F1 and OA scores. The best values are highlighted in bold font. All results are expressed as percentages ($\%$).}}
	\centering
	\scalebox{0.82}{
		\renewcommand\arraystretch{1}
		\begin{tabular}{ccccc}
			\hline
			\multicolumn{2}{c}{\textbf{Swin-V2}}              & \textbf{LEVIR-CD}        & \textbf{CDD}           & \textbf{WHU-CD}        \\ 		 
			Pre-trained       & Conf.        & F1 / OA         & F1 / OA       & F1 / OA       \\ \hline
			& Conf.1               & 91.66 / 99.17   & 97.41 / 99.36 & 95.32 / 99.61 \\		
			\checkmark        & Conf.1               & 91.77 / 99.17   & 97.46 / 99.38 
			& \textbf{95.50} / \textbf{99.62} \\			
			\checkmark        & Conf.2               & \textbf{91.85} / \textbf{99.18}   
			& \textbf{97.58} / \textbf{99.41} & 95.34 / 99.61 \\			
			\checkmark        & Conf.3               & 91.79 / 99.18   & 97.42 / 99.37 & 95.26 / 99.60 \\ \hline	
		\end{tabular}
	}
	\label{tab4}
\end{table}

\subsubsection{Ablation Study of Mixed Feature Pyramid}
we propose MFP using the combination of GT, Res2Net-Conv and SK-Conv. In the left subtable of Table \ref{tab5}, we present a detailed ablation analysis for these three modules with FC-Siam-Diff and our model as the base lines on the LEVIR-CD dataset. In the right subtable of Table \ref{tab5}, we further test the plug and play performance of MFP based on the other four models. The data of the left subtable supports the detection role of each module and the improvement of overall performance using MFP. Specifically, MFP improves F1 scores by 0.70$\%$ and 0.215$\%$ on FC-Siam-Diff and our model (our model has a high complexity), respectively. Through the analysis of the right subtable, FC-Siam-Conc, IFNet, SNUNet-CD, and FCCDN achieve the improvements of 0.77$\%$, 0.36$\%$, 0.07$\%$, and 0.46$\%$ in F1 score, respectively. The experiments on the longitudinal dissection and horizontal promotion support that MFP effectively improves the performance of change detection models by providing the inter-layer interaction information and intra-layer multi-scale information.
\begin{figure*}[t]
	\vspace{-0.4cm}
	\centering
	\includegraphics[width=0.95\linewidth]{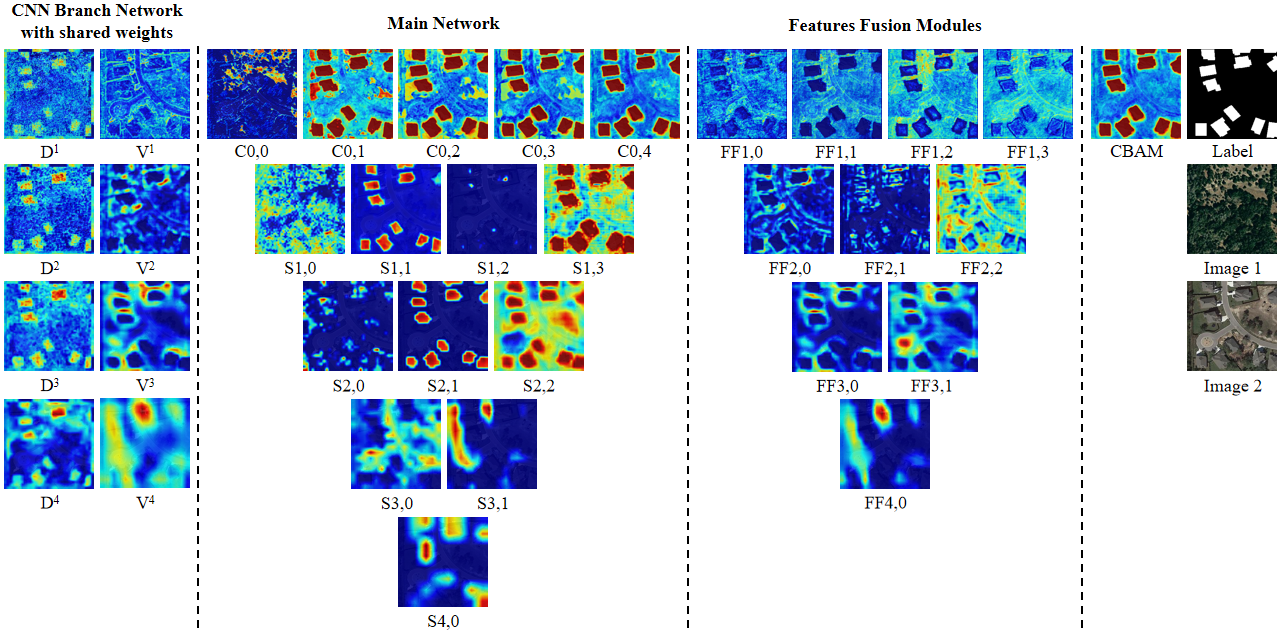}
	\caption{The example of network visualization.}
	\label{fig9}
\end{figure*}

\subsubsection{Ablation Study of SSL Strategy}
We compare the performance gains brought by guiding CNN branch network under the three strategies of unsupervised, supervised and self-supervised, as shown in Table \ref{tab6}. On the LEVIR-CD, CDD and WHU-CD datasets, self-supervised strategy achieves the best results. The reason for this phenomenon is that self-supervised learning provides the deep features with semantic information for change detection tasks \cite{chen2022fccdn}.
\begin{table}[h!]
	\caption{\centering{The ablation studies for the supervised strategy of CNN branch network on the three datasets. We report F1 and OA scores. The best values are highlighted in bold font. All results are expressed as percentages ($\%$).}}
	\centering
	\scalebox{0.85}{
		\renewcommand\arraystretch{1}
		\begin{tabular}{cccc}
			\hline
			\multirow{2}{*}{\textbf{Strategy}}     & \textbf{LEVIR-CD}        & \textbf{CDD}           & \textbf{WHU-CD}        \\ 		 
			& F1 / OA         & F1 / OA       & F1 / OA       \\ \hline
			Unsupervised                    & 91.97 / 99.19   & 97.66 / 99.43 & 95.10 / 99.59 \\			
			Supervised                      & 91.97 / 99.19   & 97.68 / 99.43 & 95.39 / 99.62 \\			
			Self-supervised                 & \textbf{92.15} / \textbf{99.21} & \textbf{97.71} / \textbf{99.44} 
			& \textbf{95.69} / \textbf{99.64} \\ \hline	
		\end{tabular}
	}
	\label{tab6}
\end{table}
\vspace{-0.1cm}

\subsection{Network Visualization}
To further elucidate the practical effect of each network module, we conduct the analysis of network visualization. As depicted in Fig. \ref{fig9}, we broadly segment the network into three components: CNN branch network, main network, and feature fusion modules. Given that CNN branch network is a dual-branch network with shared weights, we only present the activation maps for one branch.
It is distinctly seen that the encoder layers furnish attention to the low-level detail information for main network. Conversely, main network provides the structured abstract features. Feature fusion modules concentrate on detailing the differential specifics of remote sensing objects while preserving structural information.
The amalgamation and retention of these two types of information are pivotal for change detection in VHRRS images. By contrasting the label and change activation map acquired through CBAM, it becomes evident that our model exhibits the robustness in the intricate scene and different lighting condition.

\section{Conclusion}
\label{section5}
We propose an end-to-end parallel compounded network, SwinV2DNet, where the densely connected main network of Swin V2 blocks provides the change relationship features, and the CNN branch network provides the pre-changed and post-changed features for change detection. This network gleans global information via transformer while capturing precise low-level details through the encoder of CNN. The joint efforts of change relationship features, pre-changed, and post-changed features rectify the limitations of previous change detection networks reliant on either early fusion or late fusion. Furthermore, we propose a plug-and-play MFP that imparts the interlayer interaction information and intralayer multi-scale information. The experiments validate the efficacy of MFP in other change detection networks as well. We employ a SSL strategy to guide our CNN branch to provide the encoder semantic features for the main network.
We achieve the stably convergent and elegantly efficient performance on the four commonly used public datasets. Importantly, our proposed model demonstrates the advanced abilities for capturing structured, small, and edge objects. This model also shows the robust capabilities in terms of anti-interference and change discrimination, especially when the illumination difference is large or the background and targets are cluttered and similar.
In the future work, we will focus on the realization of transformer lightweight for better practicability.

{\footnotesize
\bibliographystyle{IEEEtran}
\bibliography{introduction}}

\begin{thebibliography}{10}
\providecommand{\url}[1]{#1}
\csname url@samestyle\endcsname
\providecommand{\newblock}{\relax}
\providecommand{\bibinfo}[2]{#2}
\providecommand{\BIBentrySTDinterwordspacing}{\spaceskip=0pt\relax}
\providecommand{\BIBentryALTinterwordstretchfactor}{4}
\providecommand{\BIBentryALTinterwordspacing}{\spaceskip=\fontdimen2\font plus
\BIBentryALTinterwordstretchfactor\fontdimen3\font minus
  \fontdimen4\font\relax}
\providecommand{\BIBforeignlanguage}[2]{{%
\expandafter\ifx\csname l@#1\endcsname\relax
\typeout{** WARNING: IEEEtran.bst: No hyphenation pattern has been}%
\typeout{** loaded for the language `#1'. Using the pattern for}%
\typeout{** the default language instead.}%
\else
\language=\csname l@#1\endcsname
\fi
#2}}
\providecommand{\BIBdecl}{\relax}
\BIBdecl

\bibitem{hussain2013change}
M.~Hussain, D.~Chen, A.~Cheng, H.~Wei, and D.~Stanley, ``Change detection from
  remotely sensed images: From pixel-based to object-based approaches,''
  \emph{ISPRS Journal of photogrammetry and remote sensing}, vol.~80, pp.
  91--106, 2013.

\bibitem{zhang2021moving}
J.~Zhang, X.~Jia, J.~Hu, and K.~Tan, ``Moving vehicle detection for remote
  sensing video surveillance with nonstationary satellite platform,''
  \emph{IEEE Transactions on Pattern Analysis and Machine Intelligence},
  vol.~44, no.~9, pp. 5185--5198, 2021.

\bibitem{bruzzone2009domain}
L.~Bruzzone and M.~Marconcini, ``Domain adaptation problems: A dasvm
  classification technique and a circular validation strategy,'' \emph{IEEE
  transactions on pattern analysis and machine intelligence}, vol.~32, no.~5,
  pp. 770--787, 2009.

\bibitem{benedek2011building}
C.~Benedek, X.~Descombes, and J.~Zerubia, ``Building development monitoring in
  multitemporal remotely sensed image pairs with stochastic birth-death
  dynamics,'' \emph{IEEE Transactions on Pattern Analysis and Machine
  Intelligence}, vol.~34, no.~1, pp. 33--50, 2011.

\bibitem{lu2019landslide}
P.~Lu, Y.~Qin, Z.~Li, A.~C. Mondini, and N.~Casagli, ``Landslide mapping from
  multi-sensor data through improved change detection-based markov random
  field,'' \emph{Remote Sensing of Environment}, vol. 231, p. 111235, 2019.

\bibitem{jin2017land}
S.~Jin, L.~Yang, Z.~Zhu, and C.~Homer, ``A land cover change detection and
  classification protocol for updating alaska nlcd 2001 to 2011,'' \emph{Remote
  Sensing of Environment}, vol. 195, pp. 44--55, 2017.

\bibitem{zhu2014continuous}
Z.~Zhu and C.~E. Woodcock, ``Continuous change detection and classification of
  land cover using all available landsat data,'' \emph{Remote sensing of
  Environment}, vol. 144, pp. 152--171, 2014.

\bibitem{lv2021land}
Z.~Lv, T.~Liu, J.~A. Benediktsson, and N.~Falco, ``Land cover change detection
  techniques: Very-high-resolution optical images: A review,'' \emph{IEEE
  Geoscience and Remote Sensing Magazine}, vol.~10, no.~1, pp. 44--63, 2021.

\bibitem{liangpei2017advance}
Z.~Liangpei and W.~Chen, ``Advance and future development of change detection
  for multi-temporal remote sensing imagery,'' \emph{Acta Geodaetica et
  Cartographica Sinica}, vol.~46, no.~10, p. 1447, 2017.

\bibitem{zhuang2016strategies}
H.~Zhuang, K.~Deng, H.~Fan, and M.~Yu, ``Strategies combining spectral angle
  mapper and change vector analysis to unsupervised change detection in
  multispectral images,'' \emph{IEEE Geoscience and Remote Sensing Letters},
  vol.~13, no.~5, pp. 681--685, 2016.

\bibitem{malila1980change}
W.~A. Malila, ``Change vector analysis: An approach for detecting forest
  changes with landsat,'' in \emph{LARS symposia}, 1980, p. 385.

\bibitem{bovolo2006theoretical}
F.~Bovolo and L.~Bruzzone, ``A theoretical framework for unsupervised change
  detection based on change vector analysis in the polar domain,'' \emph{IEEE
  Transactions on Geoscience and Remote Sensing}, vol.~45, no.~1, pp. 218--236,
  2006.

\bibitem{zhang2007remote}
J.~Zhang and Y.~Zhang, ``Remote sensing research issues of the national land
  use change program of china,'' \emph{ISPRS Journal of Photogrammetry and
  Remote Sensing}, vol.~62, no.~6, pp. 461--472, 2007.

\bibitem{zhong2006multi}
J.~Zhong and R.~Wang, ``Multi-temporal remote sensing change detection based on
  independent component analysis,'' \emph{International Journal of Remote
  Sensing}, vol.~27, no.~10, pp. 2055--2061, 2006.

\bibitem{xian2010updating}
G.~Xian and C.~Homer, ``Updating the 2001 national land cover database
  impervious surface products to 2006 using landsat imagery change detection
  methods,'' \emph{Remote sensing of environment}, vol. 114, no.~8, pp.
  1676--1686, 2010.

\bibitem{wu2013slow}
C.~Wu, B.~Du, and L.~Zhang, ``Slow feature analysis for change detection in
  multispectral imagery,'' \emph{IEEE Transactions on Geoscience and Remote
  Sensing}, vol.~52, no.~5, pp. 2858--2874, 2013.

\bibitem{gil2016description}
J.~L. Gil-Yepes, L.~A. Ruiz, J.~A. Recio, {\'A}.~Balaguer-Beser, and
  T.~Hermosilla, ``Description and validation of a new set of object-based
  temporal geostatistical features for land-use/land-cover change detection,''
  \emph{ISPRS Journal of Photogrammetry and Remote Sensing}, vol. 121, pp.
  77--91, 2016.

\bibitem{daudt2018fully}
R.~C. Daudt, B.~Le~Saux, and A.~Boulch, ``Fully convolutional siamese networks
  for change detection,'' in \emph{2018 25th IEEE International Conference on
  Image Processing (ICIP)}.\hskip 1em plus 0.5em minus 0.4em\relax IEEE, 2018,
  pp. 4063--4067.

\bibitem{zheng2022learning}
D.~Zheng, Z.~Wei, Z.~Wu, and J.~Liu, ``Learning pairwise potential crfs in deep
  siamese network for change detection,'' \emph{Remote Sensing}, vol.~14,
  no.~4, p. 841, 2022.

\bibitem{zhang2022swinsunet}
C.~Zhang, L.~Wang, S.~Cheng, and Y.~Li, ``Swinsunet: Pure transformer network
  for remote sensing image change detection,'' \emph{IEEE Transactions on
  Geoscience and Remote Sensing}, vol.~60, pp. 1--13, 2022.

\bibitem{alcantarilla2018street}
P.~F. Alcantarilla, S.~Stent, G.~Ros, R.~Arroyo, and R.~Gherardi, ``Street-view
  change detection with deconvolutional networks,'' \emph{Autonomous Robots},
  vol.~42, pp. 1301--1322, 2018.

\bibitem{peng2019end}
D.~Peng, Y.~Zhang, and H.~Guan, ``End-to-end change detection for high
  resolution satellite images using improved unet++,'' \emph{Remote Sensing},
  vol.~11, no.~11, p. 1382, 2019.

\bibitem{peng2020optical}
X.~Peng, R.~Zhong, Z.~Li, and Q.~Li, ``Optical remote sensing image change
  detection based on attention mechanism and image difference,'' \emph{IEEE
  Transactions on Geoscience and Remote Sensing}, vol.~59, no.~9, pp.
  7296--7307, 2020.

\bibitem{zhan2017change}
Y.~Zhan, K.~Fu, M.~Yan, X.~Sun, H.~Wang, and X.~Qiu, ``Change detection based
  on deep siamese convolutional network for optical aerial images,'' \emph{IEEE
  Geoscience and Remote Sensing Letters}, vol.~14, no.~10, pp. 1845--1849,
  2017.

\bibitem{zhang2020deeply}
C.~Zhang, P.~Yue, D.~Tapete, L.~Jiang, B.~Shangguan, L.~Huang, and G.~Liu, ``A
  deeply supervised image fusion network for change detection in high
  resolution bi-temporal remote sensing images,'' \emph{ISPRS Journal of
  Photogrammetry and Remote Sensing}, vol. 166, pp. 183--200, 2020.

\bibitem{chen2021remote}
H.~Chen, Z.~Qi, and Z.~Shi, ``Remote sensing image change detection with
  transformers,'' \emph{IEEE Transactions on Geoscience and Remote Sensing},
  vol.~60, pp. 1--14, 2021.

\bibitem{shi2021deeply}
Q.~Shi, M.~Liu, S.~Li, X.~Liu, F.~Wang, and L.~Zhang, ``A deeply supervised
  attention metric-based network and an open aerial image dataset for remote
  sensing change detection,'' \emph{IEEE transactions on geoscience and remote
  sensing}, vol.~60, pp. 1--16, 2021.

\bibitem{fang2021snunet}
S.~Fang, K.~Li, J.~Shao, and Z.~Li, ``Snunet-cd: A densely connected siamese
  network for change detection of vhr images,'' \emph{IEEE Geoscience and
  Remote Sensing Letters}, vol.~19, pp. 1--5, 2021.

\bibitem{zhang2023asymmetric}
X.~Zhang, S.~Cheng, L.~Wang, and H.~Li, ``Asymmetric cross-attention
  hierarchical network based on cnn and transformer for bitemporal remote
  sensing images change detection,'' \emph{IEEE Transactions on Geoscience and
  Remote Sensing}, vol.~61, pp. 1--15, 2023.

\bibitem{feng2023change}
Y.~Feng, J.~Jiang, H.~Xu, and J.~Zheng, ``Change detection on remote sensing
  images using dual-branch multilevel intertemporal network,'' \emph{IEEE
  Transactions on Geoscience and Remote Sensing}, vol.~61, pp. 1--15, 2023.

\bibitem{hu2018squeeze}
J.~Hu, L.~Shen, and G.~Sun, ``Squeeze-and-excitation networks,'' in
  \emph{Proceedings of the IEEE conference on computer vision and pattern
  recognition}, 2018, pp. 7132--7141.

\bibitem{wang2020eca}
Q.~Wang, B.~Wu, P.~Zhu, P.~Li, W.~Zuo, and Q.~Hu, ``Eca-net: Efficient channel
  attention for deep convolutional neural networks,'' in \emph{Proceedings of
  the IEEE/CVF conference on computer vision and pattern recognition}, 2020,
  pp. 11\,534--11\,542.

\bibitem{woo2018cbam}
S.~Woo, J.~Park, J.-Y. Lee, and I.~S. Kweon, ``Cbam: Convolutional block
  attention module,'' in \emph{Proceedings of the European conference on
  computer vision (ECCV)}, 2018, pp. 3--19.

\bibitem{szegedy2015going}
C.~Szegedy, W.~Liu, Y.~Jia, P.~Sermanet, S.~Reed, D.~Anguelov, D.~Erhan,
  V.~Vanhoucke, and A.~Rabinovich, ``Going deeper with convolutions,'' in
  \emph{Proceedings of the IEEE conference on computer vision and pattern
  recognition}, 2015, pp. 1--9.

\bibitem{yu2015multi}
F.~Yu and V.~Koltun, ``Multi-scale context aggregation by dilated
  convolutions,'' \emph{arXiv preprint arXiv:1511.07122}, 2015.

\bibitem{gao2019res2net}
S.-H. Gao, M.-M. Cheng, K.~Zhao, X.-Y. Zhang, M.-H. Yang, and P.~Torr,
  ``Res2net: A new multi-scale backbone architecture,'' \emph{IEEE transactions
  on pattern analysis and machine intelligence}, vol.~43, no.~2, pp. 652--662,
  2019.

\bibitem{li2019selective}
X.~Li, W.~Wang, X.~Hu, and J.~Yang, ``Selective kernel networks,'' in
  \emph{Proceedings of the IEEE/CVF conference on computer vision and pattern
  recognition}, 2019, pp. 510--519.

\bibitem{hou2019w}
B.~Hou, Q.~Liu, H.~Wang, and Y.~Wang, ``From w-net to cdgan: Bitemporal change
  detection via deep learning techniques,'' \emph{IEEE Transactions on
  Geoscience and Remote Sensing}, vol.~58, no.~3, pp. 1790--1802, 2019.

\bibitem{zhao2019incorporating}
W.~Zhao, L.~Mou, J.~Chen, Y.~Bo, and W.~J. Emery, ``Incorporating metric
  learning and adversarial network for seasonal invariant change detection,''
  \emph{IEEE Transactions on Geoscience and Remote Sensing}, vol.~58, no.~4,
  pp. 2720--2731, 2019.

\bibitem{chen2022fccdn}
P.~Chen, B.~Zhang, D.~Hong, Z.~Chen, X.~Yang, and B.~Li, ``Fccdn: Feature
  constraint network for vhr image change detection,'' \emph{ISPRS Journal of
  Photogrammetry and Remote Sensing}, vol. 187, pp. 101--119, 2022.

\bibitem{9874899}
H.~Chen, W.~Li, S.~Chen, and Z.~Shi, ``Semantic-aware dense representation
  learning for remote sensing image change detection,'' \emph{IEEE Transactions
  on Geoscience and Remote Sensing}, vol.~60, pp. 1--18, 2022.

\bibitem{10109662}
Y.~Zhang, Y.~Zhao, Y.~Dong, and B.~Du, ``Self-supervised pretraining via
  multimodality images with transformer for change detection,'' \emph{IEEE
  Transactions on Geoscience and Remote Sensing}, vol.~61, pp. 1--11, 2023.

\bibitem{vaswani2017attention}
A.~Vaswani, N.~Shazeer, N.~Parmar, J.~Uszkoreit, L.~Jones, A.~N. Gomez,
  {\L}.~Kaiser, and I.~Polosukhin, ``Attention is all you need,''
  \emph{Advances in neural information processing systems}, vol.~30, 2017.

\bibitem{dosovitskiy2020image}
A.~Dosovitskiy, L.~Beyer, A.~Kolesnikov, D.~Weissenborn, X.~Zhai,
  T.~Unterthiner, M.~Dehghani, M.~Minderer, G.~Heigold, S.~Gelly \emph{et~al.},
  ``An image is worth 16x16 words: Transformers for image recognition at
  scale,'' \emph{arXiv preprint arXiv:2010.11929}, 2020.

\bibitem{liu2021swin}
Z.~Liu, Y.~Lin, Y.~Cao, H.~Hu, Y.~Wei, Z.~Zhang, S.~Lin, and B.~Guo, ``Swin
  transformer: Hierarchical vision transformer using shifted windows,'' in
  \emph{Proceedings of the IEEE/CVF international conference on computer
  vision}, 2021, pp. 10\,012--10\,022.

\bibitem{liu2022swin}
Z.~Liu, H.~Hu, Y.~Lin, Z.~Yao, Z.~Xie, Y.~Wei, J.~Ning, Y.~Cao, Z.~Zhang,
  L.~Dong \emph{et~al.}, ``Swin transformer v2: Scaling up capacity and
  resolution,'' in \emph{Proceedings of the IEEE/CVF conference on computer
  vision and pattern recognition}, 2022, pp. 12\,009--12\,019.

\bibitem{hong2021spectralformer}
D.~Hong, Z.~Han, J.~Yao, L.~Gao, B.~Zhang, A.~Plaza, and J.~Chanussot,
  ``Spectralformer: Rethinking hyperspectral image classification with
  transformers,'' \emph{IEEE Transactions on Geoscience and Remote Sensing},
  vol.~60, pp. 1--15, 2021.

\bibitem{sun2022spectral}
L.~Sun, G.~Zhao, Y.~Zheng, and Z.~Wu, ``Spectral--spatial feature tokenization
  transformer for hyperspectral image classification,'' \emph{IEEE Transactions
  on Geoscience and Remote Sensing}, vol.~60, pp. 1--14, 2022.

\bibitem{gao2023adaptive}
L.~Gao, B.~Liu, P.~Fu, and M.~Xu, ``Adaptive spatial tokenization transformer
  for salient object detection in optical remote sensing images,'' \emph{IEEE
  Transactions on Geoscience and Remote Sensing}, vol.~61, pp. 1--15, 2023.

\bibitem{he2022swin}
X.~He, Y.~Zhou, J.~Zhao, D.~Zhang, R.~Yao, and Y.~Xue, ``Swin transformer
  embedding unet for remote sensing image semantic segmentation,'' \emph{IEEE
  Transactions on Geoscience and Remote Sensing}, vol.~60, pp. 1--15, 2022.

\bibitem{li2022transunetcd}
Q.~Li, R.~Zhong, X.~Du, and Y.~Du, ``Transunetcd: A hybrid transformer network
  for change detection in optical remote-sensing images,'' \emph{IEEE
  Transactions on Geoscience and Remote Sensing}, vol.~60, pp. 1--19, 2022.

\bibitem{feng2022icif}
Y.~Feng, H.~Xu, J.~Jiang, H.~Liu, and J.~Zheng, ``Icif-net: Intra-scale
  cross-interaction and inter-scale feature fusion network for bitemporal
  remote sensing images change detection,'' \emph{IEEE Transactions on
  Geoscience and Remote Sensing}, vol.~60, pp. 1--13, 2022.

\bibitem{He2015DeepRL}
K.~He, X.~Zhang, S.~Ren, and J.~Sun, ``Deep residual learning for image
  recognition,'' \emph{2016 IEEE Conference on Computer Vision and Pattern
  Recognition (CVPR)}, pp. 770--778, 2015.

\bibitem{Shelhamer2014FullyCN}
E.~Shelhamer, J.~Long, and T.~Darrell, ``Fully convolutional networks for
  semantic segmentation,'' \emph{2015 IEEE Conference on Computer Vision and
  Pattern Recognition (CVPR)}, pp. 3431--3440, 2014.

\bibitem{DBLP:journals/corr/RonnebergerFB15}
\BIBentryALTinterwordspacing
O.~Ronneberger, P.~Fischer, and T.~Brox, ``U-net: Convolutional networks for
  biomedical image segmentation,'' \emph{CoRR}, vol. abs/1505.04597, 2015.
  [Online]. Available: \url{http://arxiv.org/abs/1505.04597}
\BIBentrySTDinterwordspacing

\bibitem{Zhang2020FeaturePT}
D.~Zhang, H.~Zhang, J.~Tang, M.~Wang, X.~Hua, and Q.~Sun, ``Feature pyramid
  transformer,'' \emph{ArXiv}, vol. abs/2007.09451, 2020.

\bibitem{Feng2023ChangeDO}
Y.~Feng, J.~Jiang, H.~Xu, and J.~Zheng, ``Change detection on remote sensing
  images using dual-branch multilevel intertemporal network,'' \emph{IEEE
  Transactions on Geoscience and Remote Sensing}, vol.~61, pp. 1--15, 2023.

\bibitem{Lei2023UltralightweightSF}
T.~Lei, X.~Geng, H.~Ning, Z.~Lv, M.~Gong, Y.~Jin, and A.~Nandi,
  ``Ultralightweight spatial–spectral feature cooperation network for change
  detection in remote sensing images,'' \emph{IEEE Transactions on Geoscience
  and Remote Sensing}, vol.~61, pp. 1--14, 2023.

\bibitem{devlin2018bert}
J.~Devlin, M.-W. Chang, K.~Lee, and K.~Toutanova, ``Bert: Pre-training of deep
  bidirectional transformers for language understanding,'' \emph{arXiv preprint
  arXiv:1810.04805}, 2018.

\bibitem{chen2020simple}
T.~Chen, S.~Kornblith, M.~Norouzi, and G.~Hinton, ``A simple framework for
  contrastive learning of visual representations,'' in \emph{International
  conference on machine learning}.\hskip 1em plus 0.5em minus 0.4em\relax PMLR,
  2020, pp. 1597--1607.

\bibitem{he2022masked}
K.~He, X.~Chen, S.~Xie, Y.~Li, P.~Doll{\'a}r, and R.~Girshick, ``Masked
  autoencoders are scalable vision learners,'' in \emph{Proceedings of the
  IEEE/CVF Conference on Computer Vision and Pattern Recognition}, 2022, pp.
  16\,000--16\,009.

\bibitem{chen2103self}
Y.~Chen and L.~Bruzzone, ``Self-supervised change detection in multi-view
  remote sensing images. arxiv 2021,'' \emph{arXiv preprint arXiv:2103.05969}.

\bibitem{zhang2023self}
Y.~Zhang, Y.~Zhao, Y.~Dong, and B.~Du, ``Self-supervised pre-training via
  multi-modality images with transformer for change detection,'' \emph{IEEE
  Transactions on Geoscience and Remote Sensing}, 2023.

\bibitem{simonyan2014very}
K.~Simonyan and A.~Zisserman, ``Very deep convolutional networks for
  large-scale image recognition,'' \emph{arXiv preprint arXiv:1409.1556}, 2014.

\bibitem{qin2013object}
Y.~Qin, Z.~Niu, F.~Chen, B.~Li, and Y.~Ban, ``Object-based land cover change
  detection for cross-sensor images,'' \emph{International Journal of Remote
  Sensing}, vol.~34, no.~19, pp. 6723--6737, 2013.

\bibitem{chen2020spatial}
H.~Chen and Z.~Shi, ``A spatial-temporal attention-based method and a new
  dataset for remote sensing image change detection,'' \emph{Remote Sensing},
  vol.~12, no.~10, p. 1662, 2020.

\bibitem{lebedev2018change}
M.~Lebedev, Y.~V. Vizilter, O.~Vygolov, V.~Knyaz, and A.~Y. Rubis, ``Change
  detection in remote sensing images using conditional adversarial networks.''
  \emph{International Archives of the Photogrammetry, Remote Sensing \& Spatial
  Information Sciences}, vol.~42, no.~2, 2018.

\bibitem{ji2018fully}
S.~Ji, S.~Wei, and M.~Lu, ``Fully convolutional networks for multisource
  building extraction from an open aerial and satellite imagery data set,''
  \emph{IEEE Transactions on Geoscience and Remote Sensing}, vol.~57, no.~1,
  pp. 574--586, 2018.

\bibitem{pan2021dcff}
F.~Pan, Z.~Wu, Q.~Liu, Y.~Xu, and Z.~Wei, ``Dcff-net: A densely connected
  feature fusion network for change detection in high-resolution remote sensing
  images,'' \emph{IEEE Journal of Selected Topics in Applied Earth Observations
  and Remote Sensing}, vol.~14, pp. 11\,974--11\,985, 2021.

\end{thebibliography}
\vspace{-40pt}
\begin{IEEEbiography}
	[{\includegraphics[width=1in,height=1.25in,clip,keepaspectratio]{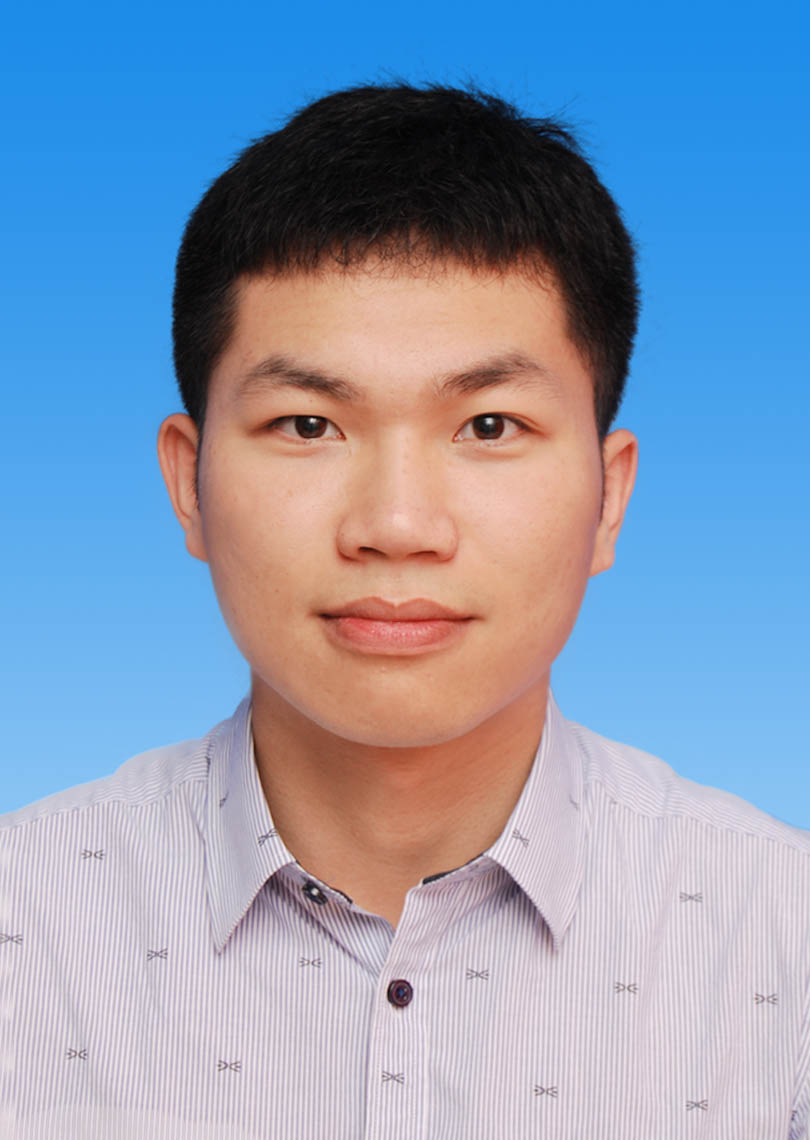}}]{Dalong Zheng}  received the B.S. degree in software engineering and M.S. degree in agricultural informatization from the Inner Mongolia Agricultural University. 
	He is currently pursuing the Ph.D. degree with the Nanjing University of Science and Technology (NJUST), Nanjing, Jiangsu, China. 
	His research interests include remote sensing image change detection, deep learning, and conditional random fields.
\end{IEEEbiography}
\vspace{-40pt}
\begin{IEEEbiography}
	[{\includegraphics[width=1in,height=1.25in,clip,keepaspectratio]{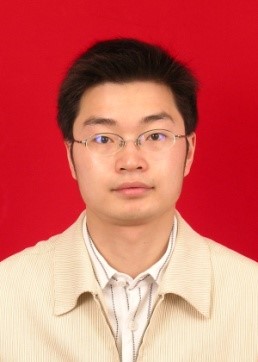}}]{Zebin Wu}
	received the B.Sc. and Ph.D. degrees in computer science and technology from the Nanjing University of Science and Technology (NJUST), in 2003 and 2007, Nanjing, China, respectively.
	
	He is currently a Professor with the School of Computer Science and Engineering, Nanjing University of Science and Technology. He was a Visiting Scholar with the GIPSA-Lab, Grenoble INP, the Université Grenoble Alpes, Grenoble, France, from August 2018 to September 2018. He was a Visiting Scholar with the Department of Mathematics, University of California at Los Angeles, Los Angeles, CA, USA, from August 2016 to September 2016 and from July 2017 to August 2017. He was a Visiting Scholar with the Hyperspectral Computing Laboratory, Department of Technology of Computers and Communications, Escuela Politécnica, University of Extremadura, Cáceres, Spain, from June 2014 to June 2015. His research interests include hyperspectral image processing, parallel computing, and big data processing.
\end{IEEEbiography}
\vspace{-40pt}
\begin{IEEEbiography}
	[{\includegraphics[width=1in,height=1.25in,clip,keepaspectratio]{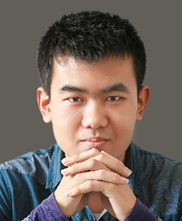}}]{Jia Liu}  received the B.S. and Ph.D. degrees in electronic engineering from the Xidian University, in 2013 and 2018, respectively. Now, he is an Associate Professor with the School of Computer Science and Engineering, Nanjing University of Science and Technology. His current research interests include computational intelligence and image understanding.
\end{IEEEbiography}
\vspace{-40pt}
\begin{IEEEbiography}
	[{\includegraphics[width=1in,height=1.25in,clip,keepaspectratio]{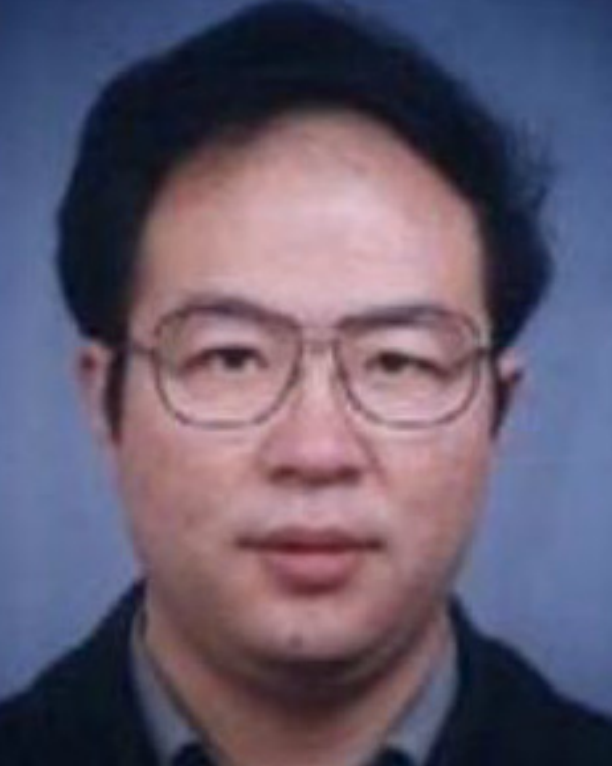}}]{Zhihui Wei} 
	was born in Jiangsu, China, in 1963. He received the B.Sc. and M.Sc. degrees in applied	mathematics and the Ph.D. degree in communication and information system from the Southeast University, Nanjing, China, in 1983, 1986, and 2003, respectively.
	
	He is currently a Professor and a Doctoral Supervisor with the Nanjing University of Science and Technology (NJUST), Nanjing. His research interests include partial differential equations, mathematical image processing, multiscale analysis, sparse representation, and compressive sensing.
\end{IEEEbiography}
\end{document}